%% file: paper.tex
\documentclass[times]{sae}
\usepackage{amsmath,amssymb,amsfonts}
\usepackage{algorithmic}
\usepackage[svgnames]{xcolor}
\definecolor{SAEblue}{RGB}{1,160,233}
\raggedright

\usepackage[justification=justified, singlelinecheck=false]{caption}  
\DeclareCaptionFont{eightpt}{\fontsize{8pt}{11pt}\selectfont #1}
\usepackage{lineno}
\usepackage[utf8]{inputenc}

\usepackage{array}
\newcolumntype{L}[1]{>{\raggedright\let\newline\\\arraybackslash\hspace{0pt}}p{#1}}
\newcolumntype{C}[1]{>{\centering\let\newline\\\arraybackslash\hspace{0pt}}p{#1}}
\newcolumntype{R}[1]{>{\raggedleft\let\newline\\\arraybackslash\hspace{0pt}}p{#1}}

\usepackage{graphicx}
\usepackage{multirow}
\usepackage{mathtools}

\usepackage[colorlinks]{hyperref}

\input{glossary}

\newcommand{\ignore}[1]{}

\usepackage{nameref}
\usepackage{booktabs}

\input{macros}

\makeatletter
\def\@seccntformat#1{%
  \expandafter\csname c@#1\endcsname\c@section
  }
\makeatother

\usepackage{subcaption}
\usepackage{multimedia}

\PaperTitle{STEAM \& MoSAFE: \\SOTIF Error-and-Failure Model \& Analysis \\for AI-Enabled Driving Automation}
\usepackage{calc} 

\makeatletter 
\renewcommand\@biblabel[1]{#1. } 
\makeatother

\AddAuthor{Krzysztof Czarnecki}{Waterloo Intelligent Systems Engineering (WISE) Lab, Dept. of Electrical and Computer Engineering, University of Waterloo, Waterloo, Canada}
\AddAuthor{Hiroshi Kuwajima}{DENSO CORPORATION, Tokyo, Japan}

\begin{document}
\maketitle
\section{Abstract}
Driving Automation Systems (DAS) are subject to complex road environments and vehicle behaviors and increasingly rely on sophisticated sensors and Artificial Intelligence (AI). These properties give rise to unique safety faults stemming from specification insufficiencies and technological performance limitations, where sensors and AI introduce errors that vary in magnitude and temporal patterns, posing potential safety risks. The Safety of the Intended Functionality (SOTIF) standard emerges as a promising framework for addressing these concerns, focusing on scenario-based analysis to identify hazardous behaviors and their causes. Although the current standard provides a basic cause-and-effect model and high-level process guidance, it lacks concepts required to identify and evaluate hazardous errors, especially within the context of AI.

This paper introduces two key contributions to bridge this gap. First, it defines the SOTIF Temporal Error and Failure Model (STEAM) as a refinement of the SOTIF cause-and-effect model, offering a comprehensive system-design perspective. STEAM refines error definitions, introduces error sequences, and classifies them as error sequence patterns, providing particular relevance to systems employing advanced sensors and AI. Second, this paper proposes the Model-based SOTIF Analysis of Failures and Errors (MoSAFE) method, which allows instantiating STEAM based on system-design models by deriving hazardous error sequence patterns at module level from hazardous behaviors at vehicle level via weakest precondition reasoning. Finally, the paper presents a case study centered on an automated speed-control feature, illustrating the practical applicability of the refined model and the MoSAFE method in addressing complex safety challenges in DAS.

\input{intro}
\input{background}

\input{steam}
\input{mosafe}

\input{limitations}

\input{related}

\input{conclusions}



\bibliographystyle{ieeetr} 
\bibliography{references}


\section{Acknowledgments}
We acknowledge Dr. Rick Salay for his valuable discussions and fruitful feedback during the research development phase. Our thanks go to Maximilian Khan and Glen Tipold for their critical role in conducting the simulation experiments that validated the UBI model for the WISE ADS. We also extend our gratitude to Dr. Andrzej Wasowski and the anonymous reviewers for their constructive comments on an earlier draft of this manuscript.


\printglossary[type=\acronymtype]
\printglossary[type=main]

\end{document}

%% file: glossary.tex
\usepackage[acronym]{glossaries}
\usepackage[automake]{glossaries-extra}
\makeglossaries

\newcommand{\newglossaryentrywithacronym}[3]{
    \newglossaryentry{#1_gls}{
        name={#1},
        long={#2},
        description={#3}
    }

    \newglossaryentry{#1}{
        type=\acronymtype,
        name={#1},
        description={#2},
        first={#2 (#1)\glsadd{#1_gls}},
        firstplural={#2s (#1s)\glsadd{#1_gls}},
        see=[Glossary:]{#1_gls}
    }
}

\newacronym{ADAS}{ADAS}{Advanced Driver Assistance System}

\newacronym{ADS}{ADS}{Automated Driving System}

\newacronym{AI}{AI}{Artificial Intelligence}

\newacronym{ALARP}{ALARP}{As Low as Reasonably Practicable} 

\newacronym{DAS}{DAS}{Driving Automation System}

\newacronym{dFHL}{dFHL}{differential Floyd-Hoare logic}

\newacronym{DOF}{DOF}{Degrees of Freedom}

\newglossaryentrywithacronym{DSM}{Detailed Scenario Model}{refinement of the \gls{HLSM} to include \glspl{IRC} and \gls{DAS} elements for \gls{HEP} identification}

\newacronym{EE}{E/E}{electrical and/or electronic}

\newglossaryentrywithacronym{FN}{False Negative}{a foreground object misclassified as background}
    
\newglossaryentrywithacronym{FP}{False Positive}{background misclassified as a foreground object}

\newacronym{FT}{FT}{Fault Tree}

\newacronym{FTA}{FTA}{Fault Tree Analysis}

\newglossaryentrywithacronym{FTY}{Failure to Yield}{an \gls{HB} representing failure to yield when required otherwise}

\newacronym{GAMAB}{GAMAB}{Globalement Au Moins Aussi Bon}

\newacronym{GSN}{GSN}{Goal Structuring Notation}
    
\newglossaryentrywithacronym{HB}{Hazardous Behavior}{\gls{DAS} behavior at the vehicle-level that may lead to harm}
    
\newglossaryentrywithacronym{HBB}{Hazardous Braking}{\gls{UBI} or \gls{UIB} or \gls{UHB}}
    
\newglossaryentrywithacronym{HBP}{Hazardous Behavior Pattern}{a class of \glspl{HB}}
    
\newglossaryentrywithacronym{HBSC}{Hazardous-Behavior-Sensitive Scenario Condition}{a scenario condition that links \gls{HB} to harm}
    
\newglossaryentrywithacronym{HEP}{Hazardous Error Pattern}{a class of hazardous error sequences (or its over-approximation)}
    
\newglossaryentrywithacronym{HES}{Hazardous Error Sequence}{a sequence of errors that causes an \gls{HB}}
    
\newglossaryentrywithacronym{HLSM}{High-Level Scenario Model}{model of the driving policy and the \gls{RVE} with \glspl{HBSC} for \gls{HBP} identification and evaluation}

\newglossaryentrywithacronym{IRC}{Input-Relevant Scenario Condition}{scenario conditions that affect \gls{DAS} inputs}

\newacronym{ISCaP}{ISCaP}{Integration Safety Case for Perception}

\newacronym{MAIS}{MAIS}{Maximum Abbreviated Injury Scale}

\newacronym{MEM}{MEM}{Minimal Endogenous Mortality}

\newglossaryentrywithacronym{MoSAFE}{Model-based \gls{SOTIF} Analysis of Failures and Errors}{a model-based method for identification and evaluation of \glspl{HBP} and \glspl{HEP}}

\newacronym{ODD}{ODD}{Operational Design Domain}

\newacronym{ODE}{ODE}{Ordinary Differential Equation}

\newglossaryentrywithacronym{POV}{Principle Other Vehicle}{the main other vehicle interacting with the \gls{SV} in a scenario}

\newacronym{RSS}{RSS}{Responsibility-Sensitive Safety}

\newglossaryentrywithacronym{RVE}{Road-and-Vehicle Environment}{environment of the \gls{DAS}, consisting of the \gls{SV} and its road environment}

\newacronym{SCM}{SCM}{Structural Causal Model}

\newacronym{SOTIF}{SOTIF}{Safety of the Intended Functionality}

\newglossaryentrywithacronym{STEAM}{\gls{SOTIF} Temporal Error and Failure Model}{a refinement of \gls{SOTIF} cause-and-effect model with error patterns and scenario-condition categories}

\newglossaryentrywithacronym{SV}{Subject Vehicle}{a vehicle controlled by the \gls{DAS} under analysis}

\newglossaryentrywithacronym{TP}{True Positive}{correct detection}

\newglossaryentrywithacronym{UA}{Unintended Acceleration}{acceleration by the \gls{SV} when the intended behavior is to maintain speed}

\newglossaryentrywithacronym{UBI}{Unintended Braking Interruption}{lack of braking when the intended behavior is to brake}

\newglossaryentrywithacronym{UHB}{Unintended Hard Braking}{hard braking when the intended behavior is no braking or braking at lower intensity}

\newglossaryentrywithacronym{UIB}{Unintended Insufficient Braking}{braking at lower intensity than intended}

\newglossaryentrywithacronym{WPP}{Weakest Precondition Pattern}{\gls{HEP} consisting of all \glspl{HES} causing another \gls{HEP} or \gls{HBP}}

%% file: macros.tex
\DeclarePairedDelimiter\ceil{\lceil}{\rceil}
\DeclarePairedDelimiter\floor{\lfloor}{\rfloor}

\newtheorem{mymethod}{Method}
\newtheorem{mydefinition}{Definition}
\newtheorem{myobservation}{Observation}
\newtheorem{myremark}{Remark}
\newtheorem{myproposition}{Proposition}
\newtheorem{myclaim}{Claim}
\newtheorem{mylemma}{Lemma}
\newtheorem{mycorollary}{Corollary}
\newtheorem{myexample}{Example}
\newtheorem{myexamples}{Examples}
\newtheorem{myalgorithm}{Algorithm}
\newtheorem{myconstruction}{Construction}
\newtheorem{myrule}{Rule}

\newcommand{\bolddot}{\hspace{-1.5mm}\textbf{.}\ \  }

\newcommand{\BT}{\begin{theorem}}
\newcommand{\ET}{\end{theorem}}
\newcommand{\BCR}{\begin{mycorollary}\bolddot}
\newcommand{\ECR}{\end{mycorollary}}
\newcommand{\BPR}{\begin{myproposition}}
\newcommand{\EPR}{\end{myproposition}}
\newcommand{\BL}{\begin{mylemma}}
\newcommand{\EL}{\end{mylemma}}
\newcommand{\BCM}{\begin{myclaim}}
\newcommand{\ECM}{\end{myclaim}}

\newcommand{\BD}{\begin{mydefinition}}
\newcommand{\ED}{\end{mydefinition}}
\newcommand{\BPF}{\begin{proof}}
\newcommand{\EPF}{\end{proof}}
\newcommand{\BEX}{\begin{myexample}}
\newcommand{\EEX}{\end{myexample}}
\newcommand{\BEXS}{\begin{myexamples}}
\newcommand{\EEXS}{\end{myexamples}}
\newcommand{\BOB}{\begin{myobservation}}
\newcommand{\EOB}{\end{myobservation}}
\newcommand{\BR}{\begin{myremark}}
\newcommand{\ER}{\end{myremark}}
\newcommand{\BAL}{\begin{myalgorithm}}
\newcommand{\EAL}{\end{myalgorithm}}
\newcommand{\BAM}{\begin{mymethod}}
\newcommand{\EAM}{\end{mymethod}}

\newcommand{\BCO}{\begin{myconstruction}}
\newcommand{\ECO}{\end{myconstruction}}
\newcommand{\BRule}{\begin{myrule}}
\newcommand{\ERule}{\end{myrule}}

%% file: intro.tex
\section{Introduction}
\label{sec:intro}

With the rapid proliferation of \gls{DAS} in vehicles, such as \glsxtrfullpl{ADAS} and \glsxtrfullpl{ADS}, assuring their safety becomes paramount. This paper addresses the challenges of safety assurance of DAS that are subject to complex road environments and vehicle behaviors, exacerbated by the growing reliance on sophisticated sensors and \gls{AI}. Such challenges give rise to unique safety faults stemming from specification insufficiencies and technological performance limitations, where sensors and \gls{AI} introduce errors that vary in magnitude and temporal patterns, posing potential safety risks. For example, an AI-based object detector may experience a \gls{FN} detection in a sensor frame. While a singular \gls{FN} may not cause any safety risk, this error repeating multiple times while approaching an obstacle may cause a collision. The \gls{SOTIF} standard ISO 21448:2022\cite{ISO21448}  emerges as a promising framework for addressing these concerns, focusing on scenario-based analysis to identify hazardous behaviors and their causes. Although the current standard provides a basic cause-and-effect model linking specification insufficiencies and technological performance limitations to hazardous behaviors, it lacks concepts required to adequately specify hazardous errors, especially within the context of \gls{AI}, such as the sequences of \glspl{FN} that may lead to a crash. Further, SOTIF suggests the analysis of system architecture to identify potential functional insufficiencies, but it does not provide any concrete method to do so. It also does not give detailed guidance on severity evaluation of the identified hazards.

This paper introduces two key contributions to bridge this gap. First, it defines the \gls{STEAM}, offering a comprehensive system-design perspective. \gls{STEAM} refines error definitions, introduces error sequences, and classifies them as error sequence patterns, providing particular relevance to systems employing advanced sensors and \gls{AI}. Furthermore, it categorizes scenario conditions based on their role in the causal chain, enabling a gradual refinement of scenario and system behavior models for safety analysis. Second, this paper proposes the \gls{MoSAFE} method, building upon the \gls{STEAM}. \gls{MoSAFE} leverages system design models, scenarios, and harmful events to derive scenario-specific causal error and failure models. This is achieved by adapting \acrfullpl{FT} to incorporate error sequence patterns as nodes. The approach enables the derivation of hazardous error sequence patterns from hazardous behaviors at vehicle level using weakest precondition analysis, allowing for probabilistic analysis of error occurrence, particularly in the context of sensors and \gls{AI}. To enhance tractability, conservative approximation techniques are employed. Finally, the paper presents a case study centered on an automated braking feature, illustrating the practical applicability of the refined model and the \gls{MoSAFE} method in addressing complex safety challenges in \gls{DAS}.

In summary, this paper makes the following contributions:
\begin{enumerate}
    \item \gls{STEAM} refines the \gls{SOTIF} cause-and-effect model by adding the concept of (i) hazardous error sequences to recognize the spatio-temporal nature of hazardous errors; and (ii) \glspl{HBP} at vehicle level and (iii) \glspl{HEP} at element level as a means to specify classes of hazardous behaviors and hazardous error sequences, respectively. It also categorizes scenario conditions based on their role in linking hazardous behaviors to harm and their effects on DAS inputs.
    \item \gls{MoSAFE} is a model-based method to identify \glspl{HBP} and \glspl{HEP} and evaluate their severity and likelihood as part of SOTIF analyses in Clause 6 and Clause 7 of ISO 21448:2022~\cite{ISO21448}. \gls{MoSAFE} relies on building scenario-specific models of the \gls{DAS} and its road-and-vehicle environment, which are instrumented to inject deviations from the intended behavior. \gls{MoSAFE} uses these models to identify \glspl{HBP} and derive \glspl{HEP} from \glspl{HBP} as weakest preconditions (or their over-approximations). It additionally captures the causal links among the \glspl{HBP} and \glspl{HEP} in a novel form of an \gls{FT} with temporal patterns as nodes. The \gls{FT} can also express over-approximations using implication arrows. Finally, \gls{MoSAFE} allows deriving safety requirements on the performance of AI-based components as upper bounds on \gls{HEP} occurrence rates, and  \gls{MoSAFE} results can be used as evidence in a safety case.
\end{enumerate}

%% file: background.tex
\section{Background: SOTIF Cause-and-Effect Model and Assurance Process}
\label{sec:background}

\begin{figure*}
    \centering
    \includegraphics[width=1\textwidth]{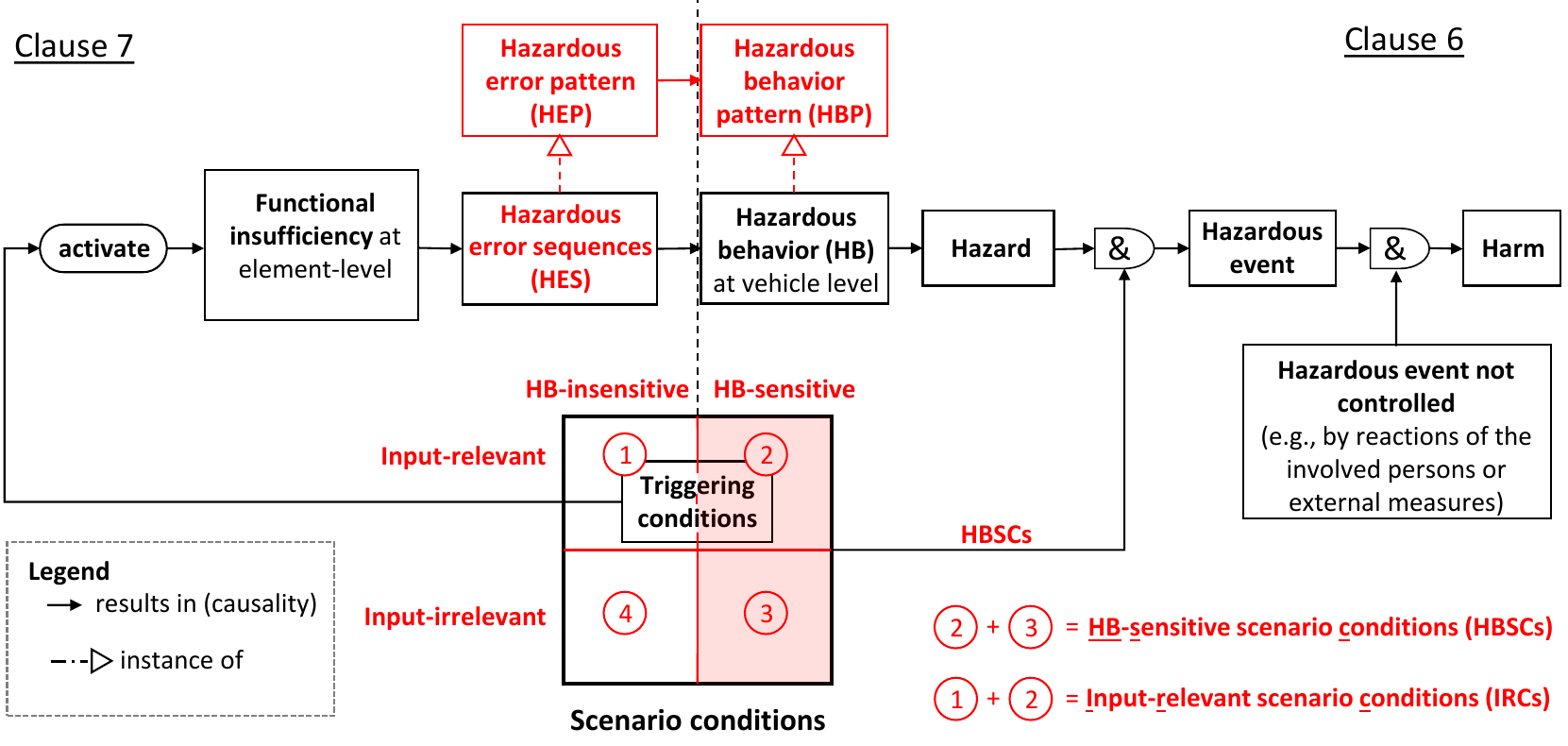}
    \caption{SOTIF cause-and-effect model (based on Figs.~3B and 4 in \cite{ISO21448}), including new elements of SOTIF Temporal Error and Failure Model (STEAM) in red}
    \label{fig:steam}
\end{figure*}

ISO 21448:2022~\cite{ISO21448} is an international standard providing guidance on assuring the safety of the intended functionality (\gls{SOTIF}) of \glsxtrfull{EE} systems (including software), especially emergency intervention systems and \gls{DAS} at SAE levels 1 to 5~\cite{J3016}. \gls{SOTIF} is defined as the absence of unreasonable risk due to a hazard caused by \emph{functional insufficiencies}, which are (i) \emph{insufficiencies of specification} of the intended functionality or (ii) \emph{performance insufficiencies}, both at the vehicle level or the level of the \gls{EE} elements implementing the \gls{DAS}. \gls{SOTIF} complements ISO 26262:2018~\cite{ISO26262}, which focuses on functional safety assurance (FuSA), that is, assuring the absence of unreasonable risk due to a hazard caused by deviating from the specified behavior. While assuring the absence of unreasonable risk due to a hazard caused by functional insufficiencies of components is the subject of both standards, functional insufficiencies of \gls{AI} components are currently insufficiently covered by ISO 26262:2018~\cite{salay2018using}, but are explicitly in scope of \gls{SOTIF}.

Fig.~\ref{fig:steam} summarizes the cause-and-effect model that underlies \gls{SOTIF} at element level (ignore the red elements for now). \gls{SOTIF} focuses on hazards that result from functional insufficiencies of the \gls{DAS}, which consist of insufficiencies of specification and performance insufficiencies at the vehicle and element level, where ``element'' refers to one or more hardware parts and software units of the \gls{DAS}. Fig.~\ref{fig:steam} explicitly shows functional insufficiencies at the element level, but functional insufficiencies at the vehicle level are also considered in our analysis, as will be explained. An example of an insufficiency of specification at the vehicle level is an incorrectly specified vehicle behavior to be implemented by the \gls{DAS}, such as an inadequate braking level in a given scenario. An example of an insufficiency of specification at the element level is the detection range of the object detector that is inadequately selected for the target \glsxtrfull{ODD}. An example of a performance insufficiency at the element level is an insufficient obstacle detection rate by the object detector. These insufficiencies can cause \gls{HB} of the \gls{SV}, such as unintended lack of braking, which may lead to \emph{hazards}, defined as potential sources of harm, such as the potential of colliding with an object. The realization of the hazard and its potential severity depend on the operational scenario in which the \gls{HB} transpires. In particular, the scenario may contain \emph{scenario conditions} in which the \gls{HB} can lead to harm, such as the presence of an obstacle blocking the lane ahead of the \gls{SV}. The occurrence of an \gls{HB} under such conditions is referred to as a \emph{hazardous event}. An example is the unintended lack of braking when approaching a stopped vehicle such that the lack of braking can cause a collision. The harm from a hazardous event may be avoided by proper reactions of the involved persons, including the \gls{SV} driver (for a driver assistance system) or the drivers of the other involved vehicles. For example, the collision may be avoided by the front vehicle accelerating or changing lane. The functional insufficiencies that lead to an \gls{HB} do so when activated by specific scenario conditions, which are referred to as \emph{triggering conditions}. In particular, a functional insufficiency on element level is activated by a triggering condition, which is a combination of scenario conditions that results in an hazardously erroneous output of the element, which then contributes to an \gls{HB} on vehicle level (see Fig.~\ref{fig:steam}). An example of an output error from an object detector is an \gls{FN}. The triggering condition for an \gls{FN} would include the object that is missed, but also other scenario conditions that contribute to the object not being detected, such as adverse weather conditions or an unusual appearance of the object. Another \gls{SOTIF} concept is as a \emph{reasonably foreseeable misuse}, which a usage of the \gls{DAS} in a way that was not intended by the manufacturer. It could itself be a triggering condition leading to an \gls{HB}, or it could contribute to reduced controllability of an \gls{HB}. Reasonably foreseeable misuse is outside the scope of this paper.

The \gls{SOTIF} standard defines a multi-activity assurance process to identify and eliminate hazards or reduce risks related to \gls{SOTIF}, spanning multiple document clauses. The process starts with the system specification and design, which among others specify the \gls{ODD}, use cases, the driving policy and the system design (Clause 5). The next activity is the identification and evaluation of \gls{SOTIF} hazards (Clause 6), which has three objectives: (1) identification of hazards; (2) evaluation of severity, exposure, and controllability, and (3) specification of acceptance criteria. The latter are used to determine whether the risk estimated in subsequent activities is reasonable. If this activity establishes that the severity and controllability of the identified hazard is above the lowest classes, respectively, S0 and C0, the potential \gls{SOTIF} causes of the hazard need to be analyzed (Clause 7), otherwise the risk is deemed reasonable for the system to be deployed. The identification and evaluation of functional insufficiencies and triggering conditions (Clause 7) has two objectives: (1) identification of functional insufficiencies (i.e., insufficiencies of specification and performance insufficiencies) and triggering conditions; and (2) evaluation of system response to the identified functional insufficiencies and triggering conditions. The latter objective requires estimating the likelihood of the hazards resulting from the identified functional insufficiencies and triggering conditions, so that their risk can be evaluated against the acceptance criteria from Clause 6. If the risk is deemed as reasonable, the system is subject to verification and validation (defined in Clause 9), which covers the known unsafe scenarios (Clause 10) and performs an exploration to uncover unknown unsafe scenarios (Clause 11). If at the end of any of these steps the risk does not meet the acceptance criteria, the system is modified to reduce the risk (Clause 8); otherwise, the residual risk is evaluated (Clause 12), and the system can be deployed. Additional \gls{SOTIF} activities occur during operation in order to uncover any additional potential \gls{SOTIF} issues (Clause 13), which then are cycled back into the \gls{SOTIF} process.

The focus of this paper is on Clauses 6 and 7. Clause 6 focuses on identifying and analyzing the effects of \glspl{HB}, which can be seen as hazardous vehicle-level failures, without the regard for their causes (the part of the cause-and-effect model on the right of the vertical dashed line in Fig.~\ref{fig:steam}). Clause 7 focuses on identifying and analyzing the causes of \glspl{HB}, especially the functional insufficiencies and the errors they cause at the element level (the part on the left of the vertical dashed line in Fig.~\ref{fig:steam}).

%% file: steam.tex
\section{SOTIF Temporal Error and Failure Model (STEAM)}
\label{sec:steam}

This section introduces our proposed \underline{S}OTIF \underline{T}emporal \underline{E}rror and F\underline{a}ilure \underline{M}odel (\gls{STEAM}) as a refinement of the original \gls{SOTIF} cause-and-effect model (see Fig.~\ref{fig:steam}). STEAM makes three refinements.

On the vehicle-level (Clause 6), \gls{STEAM} introduces the concept of a \emph{hazardous behavior pattern} (\gls{HBP}), which a specification of the magnitude and temporal occurrence of the deviations of the \gls{SV} behavior from its intended behavior that is hazardous under the given scenario conditions. Small deviations from the intended behavior may not be hazardous. For example, slightly reduced braking level when approaching a stopped vehicle may lead to a slight overshoot of the targeted stopping point, but it would not be hazardous if the \gls{SV} still stops well before the stopped vehicle. Similarly, a temporary deviation from the intended braking level, even if large, may still not be hazardous if it can be compensated by subsequently harder braking. 

On the element-level (Clause 7), \gls{STEAM} introduces the concept of a \emph{hazardous error sequence} (\gls{HES}), which is a temporal sequence of errors that is caused by a functional insufficiency of an element, and it causes an \gls{HB}. An error sequence is a temporal refinement of the concept of error from ISO 26262:2018, defined as a ``discrepancy between a computed, observed, or measured value or condition, and the true, specified or theoretically correct value or condition.''~\cite{ISO26262} The notion of the true or correct value depends on the \gls{DAS} function. 
Typical functions in a \gls{DAS} include perception, prediction, planning, and control~\cite{J3131-202203}.
For perception functions, error is defined wrt. ground truth. There is neither ground truth for prediction nor planning functions. Prediction should adequately reflect the distribution of possible futures. An example of a prediction error would be to miss a likely future, such as to exclude the possibility of another vehicle tuning when the turn is plausible. For planning, the planned actions can be assessed for their quality or level of deviation from a desired policy. For all three types of functions, a momentary error is not necessarily hazardous; it is typically a certain pattern of errors over time that becomes hazardous. For example, a single, momentary \gls{FN} may cause a momentary lack of braking, but such lack of braking may be compensated by subsequent harder braking. However, a persistent \gls{FN} or lack of braking for an obstacle ahead may become impossible to compensate and will cause a collision if not controlled for. A \emph{hazardous error pattern} (\gls{HEP}) is a specification of the pattern of errors in terms of their time and magnitude to cause an \gls{HB}. As such, an \gls{HEP} represents a set of concrete \glspl{HES} (see Fig.~\ref{fig:steam}), which may lead to a class of \glspl{HB}; and the latter may itself be specified by an \gls{HBP}. For example, an \gls{HEP} for \glspl{FN} could be specified as the total duration of missing detection of an obstacle during an approach scenario. This \gls{HEP} would include error sequences where the missing detection is in one continuous period or spans multiple periods during the scenario, as long as the total duration meets the \gls{HEP} specification. Whereas Fig.~\ref{fig:steam} shows a single \gls{HES} (and a single \gls{HBP}) causing an \gls{HB}, there will normally be a chain of \glspl{HES} through the system before they trigger an \gls{HB}. For example, a perception \gls{HES} may cause a prediction \gls{HES}, and the latter may cause a planning \gls{HES} leading to an \gls{HB}. Thus, \gls{STEAM} needs to be instantiated as error-and-failure causal chains across the \gls{DAS} design, where \glspl{HB} are considered as hazardous \gls{DAS} \emph{failures}, for specific combinations of \gls{DAS} design, hazards, and scenarios, as part of assurance. 

As a third refinement, \gls{STEAM} categorizes scenario conditions according to their role in the causal chain (see the four quadrants in Fig.~\ref{fig:steam}). First, scenario conditions are categorizes whether they influence the translation of an \gls{HB} into harm or whether they do not (see the horizontal dimension of the four quadrants in Fig.~\ref{fig:steam}): \emph{\gls{HB}-sensitive} scenario conditions (\glsdisp{HBSC}{HBSC}) are those in which the \gls{HB} leads to harm; conversely, \emph{\gls{HB}-insensitive} scenario conditions are those that do not influence the translation of \gls{HB} into harm. An example of an \gls{HBSC} for the \gls{HB} ``lack of braking'' is the presence of an obstacle ahead of the \gls{SV}; however, the color of the obstacle is an \gls{HB}-insensitive scenario condition. Second, scenario conditions are classified whether they affect the input into the \gls{DAS} or not (see the vertical dimension): \emph{input-relevant} scenario conditions (\glsdisp{IRC}{IRC}) are those affecting the \gls{DAS} input, including sensor inputs, vehicle-to-vehicle and vehicle-to-infrastructure messages, and pre-recorded maps; otherwise, they are \emph{input-irrelevant}. For example, assuming a \gls{DAS} with a camera sensor, both the presence and the color of an obstacle are \emph{input-relevant}, as they influence the image produced by the camera. Thus, the presence of the obstacle falls into quadrant 2, and its color falls into quadrant 1. Road friction is an example of an \gls{HBSC} that is not affecting the camera and thus is input-irrelevant and falls into quadrant 3. Triggering conditions are necessarily input-relevant, but not all input-relevant scenario conditions are triggering conditions. For example, the color of an obstacle may or may not be responsible for a particular \gls{FN}. 

The classification of scenario conditions enables a gradual refinement of scenario behavior models for safety analysis in Clause 6 and 7. The \glspl{HBSC} (quadrants 2 and 3 in Fig.~\ref{fig:steam}) and driving policy are relevant in the high-level scenario modeling that targets the severity assessment of hazardous behavior (Clause 6). \glspl{IRC} (quadrants 1 and 2) are relevant---in addition to quadrant 3---in the detailed modeling and analysis of scenarios, including the \gls{DAS} design, to uncover the causal chains through the system that trigger \glspl{HB} (Clause 7). Quadrant 2 impacts both the \gls{DAS} input and the translation of \glspl{HB} into harm. Quadrant 4 can be ignored during modeling and analysis.

%% file: mosafe.tex
\section{Model-based SOTIF Analysis of Failures and Errors (MoSAFE)}
\label{sec:mosafe}

We now describe our proposed \underline{Mo}del-based \underline{S}OTIF \underline{A}nalysis of \underline{F}ailures and \underline{E}rrors (\gls{MoSAFE}) method, which allows instantiating \gls{STEAM} based on scenarios; \gls{DAS}, \gls{SV}, and road environment models; and hazards. The method is divided into two activities, matching Clause 6 and 7, respectively. The first activity focuses on the identification and evaluation of \glsfirstplural{HBP}, which corresponds to the right side of the \gls{STEAM} in Fig.~\ref{fig:steam}. The second activity focuses on the identification and evaluation of \glsfirstplural{HEP}, which corresponds to the left side of the \gls{STEAM} in Fig.~\ref{fig:steam}. Both activities rely on modeling the \gls{DAS} and its environment, targeting a level of abstraction that is appropriate for the given activity's objective. The first activity's objective is to identify the \glspl{HEP} and evaluate their severity. For this purpose, it uses a high-level model of \gls{DAS}, being the intended driving policy, and an environment model capturing the \glspl{HBSC}. The second activity's objective is to identify the \glspl{HEP} and evaluate their likelihood. Therefore, this activity uses a more detailed design model of the \gls{DAS} and an environment model refined with the \glspl{IRC}. Together, the two activities provide a risk evaluation of the identified hazards, consisting of their severity and likelihood.

\subsection{Identification and Evaluation of Hazards (Clause 6)}
\label{subsec:clause6}

\begin{figure}
    \centering
    \includegraphics[width=0.47\textwidth]{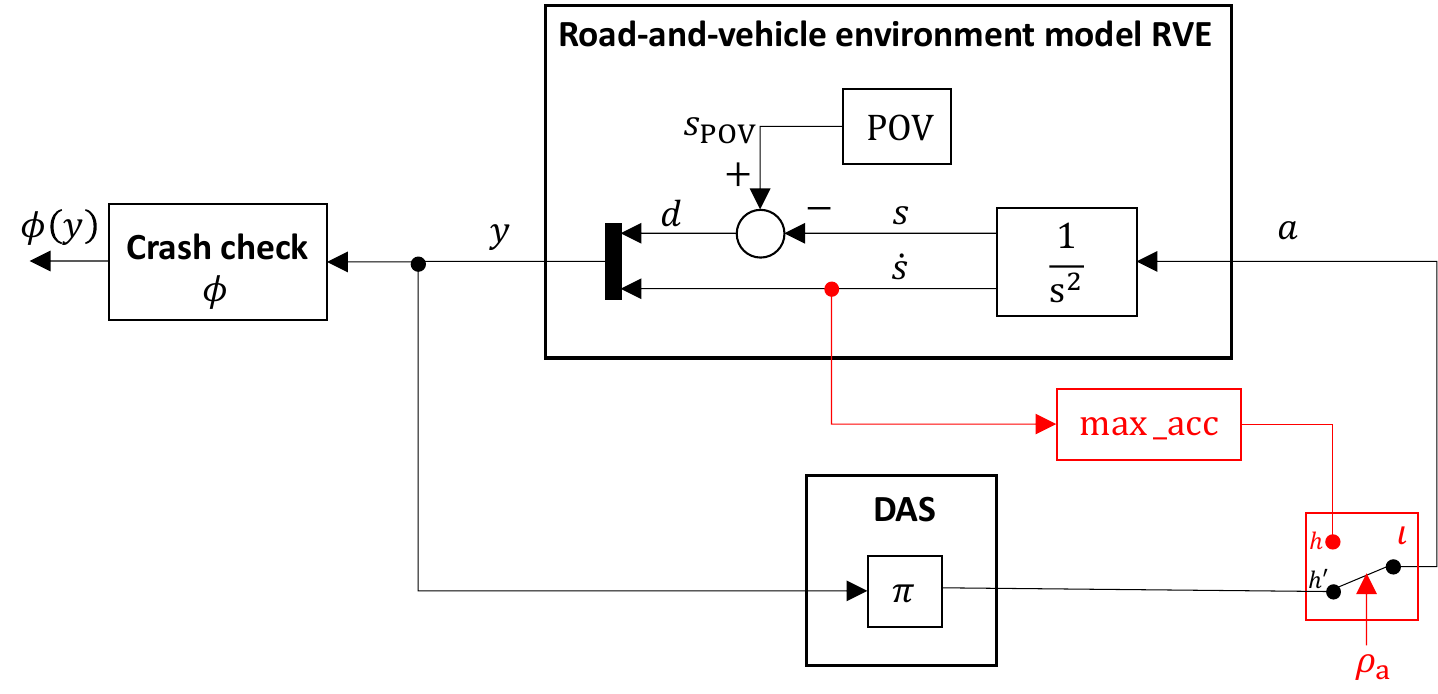}
    \caption{High-level scenario model of the \glsfirst{RVE}, capturing the \glsfirst{HBSC}, and the \gls{DAS} driving policy for ``braking for a stationary vehicle ahead'', with the \gls{HB} injection logic for \gls{UBI} marked in red}
    \label{fig:high_level_model}
\end{figure}

Clause 6 focuses on the identification and evaluation of \gls{SOTIF} hazards (the right side of Fig.~\ref{fig:steam}), and the main idea of its refinement as part of \gls{MoSAFE} is to specify them as \glspl{HBP} and evaluate their severity using high-level behavior models of the \gls{DAS} and its environment (Fig.~\ref{fig:high_level_model}).

\subsubsection{Hazard Identification}

SOTIF hazards are potential sources of harm caused by the \glspl{HB} at the vehicle level~\cite{ISO21448}, and thus their identification involves the identification of the \glspl{HB} and the \glspl{HBSC} (see the right side of Fig.~\ref{fig:steam}). The identification of \glspl{HB} involves both analyzing the safety of the specified behavior on vehicle level and the safety of deviations from the specified behavior. The deviations can be identified using Hazard and Operability Analysis (HAZOP)~\cite{HAZOP}, by instantiating guide words such as ``no'', ``more'', and ``less''. The \glspl{HBSC} are identified by systematically eliciting the operational scenarios relevant to the \gls{ODD} of the \gls{DAS} (see~\cite{ISO34502} for additional guidance).


Table~\ref{tab:hazards} lists two sample intended behaviors and the corresponding \glspl{HB} and hazards for an automated speed control feature. The feature could be part of an \gls{ADAS}, such as a full-speed range adaptive cruise control, or it could represent the longitudinal behavior aspect of an \gls{ADS}. The sample intended ``behaviors are braking for a stationary vehicle ahead'' and ``maintaining a safe distance when following a vehicle.'' The second column lists \glspl{HB}, which are hazardous deviations from the intended behavior, including \glsfirst{UBI} (i.e., \emph{no} braking when braking needed), \glsfirst{UIB} (i.e., \emph{less} braking than needed), Unintended Hard Braking (i.e., \emph{more} braking than needed), and \glsfirst{UA} (i.e., \emph{more} acceleration than needed). The third column lists the hazards resulting from the \glspl{HB}: rear-ending another vehicle or being rear-ended by another vehicle. The hazards indicate the key \glspl{HBSC} in which the given \gls{HB} may lead to harm, such as the presence of a stationary or braking vehicle ahead. These initial \glspl{HBSC} are then refined to cover the full range of operational scenarios and conditions within the \gls{ODD} of the \gls{DAS}, such as the full ranges of speed, road friction, road grade, and road curvature occurring within the \gls{ODD}.

\begin{table}
    \centering
    \begin{tabular}{|p{0.10\textwidth}|p{0.12\textwidth}|p{0.15\textwidth}|} \hline 
     \textbf{Intended longitudinal behavior} & \textbf{Longitudinal \gls{HB}} & \textbf{Hazard} \\ \hline 
     Braking for a stationary vehicle ahead   & Unintended\newline braking\newline interruption (\glsdisp{UBI}{UBI}) & Rear-end collision with the stationary vehicle \\ \cline{2-3} 
      & Unintended\newline insufficient\newline braking (\glsdisp{UIB}{UIB}) & Rear-end collision with the stationary vehicle \\ \cline{2-3} 
     & Unintended hard braking (\glsdisp{UHB}{UHB}) & Another vehicle hitting from behind \\ \hline 
     Maintaining a safe distance when following a vehicle & Unintended\newline braking\newline interruption (\glsdisp{UBI}{UBI}) & Rear-end collision with the front vehicle when it brakes  \\ \cline{2-3} 
      & Unintended\newline acceleration (\glsdisp{UA}{UA}) & Rear-end collision with the front vehicle \\ \cline{2-3} 
         & Unintended hard braking (\glsdisp{UHB}{UHB}) & Another vehicle hitting from behind \\ \hline
    \end{tabular}
    \caption{Sample intended behaviors, \glspl{HB}, and hazards related to longitudinal behavior of a \gls{DAS}}
    \label{tab:hazards}
\end{table}

\subsubsection{Hazard Evaluation and Acceptance Criteria Specification}

Hazard identification is followed by the evaluation of the severity, exposure, and controllability of the identified hazards. As recommended by the \gls{SOTIF} standard, this step leverages concepts and methods from Part 3 of ISO 26262:2018~\cite{ISO26262}, including the classification of severity (S0-S3), exposure (E0-E4), and controllability (C0-C3). For collisions, severity can be estimated using the Delta-V method~\cite{jurewicz16impact}, which maps the collision configuration and the change in velocity resulting from the collision to a range of injury severity based on crash statistics. Exposure is the estimated likelihood of the crash-relevant \glspl{HBSC} during the operation of the \gls{DAS}. Finally, controllability depends on the type of \gls{DAS} and its level of automation. Whereas \gls{ADAS} rely on the driver to intervene, an \gls{ADS}-equipped vehicle may be driverless, and the ability of other road users to control the hazardous event is often limited.

Acceptance criteria are qualitative or quantitative criteria representing the absence of an unreasonable level of risk. An example of quantitative acceptance criteria would be an upper bound on the crash rate for each severity class, possibly expressed as the mean distance travelled between crashes. Acceptance criteria may be allocated to different combinations of \glspl{HB} and \glspl{HBSC}.
For example, for an \gls{HBSC} set and its exposure $P(\text{\gls{HBSC}})$, acceptance criteria would put an upper bound on $P(\text{\gls{HB}}\mid\text{\gls{HBSC}})$, the conditional probability of \gls{HB} under \gls{HBSC}, and the probability of the hazardous scenario would be given as follows:
\begin{equation}
    P(\text{\gls{HB}},\text{\gls{HBSC}})=P(\text{\gls{HB}}\mid\text{\gls{HBSC}})P(\text{\gls{HBSC}})
\end{equation}

\subsubsection{Model-based Severity Evaluation}

The \gls{MoSAFE} method leverages high-level behavior models to evaluate the severity of the different degrees of \gls{HB} under the different \glspl{HBSC}. Although Delta-V allows evaluating the severity of harm resulting from a collision, hazard evaluation also requires mapping different degrees of \gls{HB} under different \glspl{HBSC} to crash severity. For example, the duration and timing of a braking interruption or the level of insufficient braking when approaching a stationary or braking vehicle will influence the occurrence of a collision and its severity. As part of severity evaluation, \gls{MoSAFE} uses a \gls{HLSM}, consisting of (i) the driving policy of the \gls{DAS} as its specified intended behavior and (ii) a \gls{RVE} model that captures the \glspl{HBSC}. \glspl{HLSM} are specific to the intended behavior and the \glspl{HBSC} being evaluated. In other words, ``braking for a stationary vehicle ahead'' and ``maintaining a safe distance when following a vehicle'' would each lead to a different \gls{HLSM}. Given an \gls{HLSM}, the safety of the specified behavior of the \gls{DAS} under the given \glspl{HBSC} is first evaluated. This is followed by the evaluation of \glspl{HB}, which are injected into the \gls{HLSM}. Varying levels of \glspl{HB} are then linked to crashes of different severity.

\begin{figure}[h]
    \centering
    \includegraphics[width=0.3\textwidth]{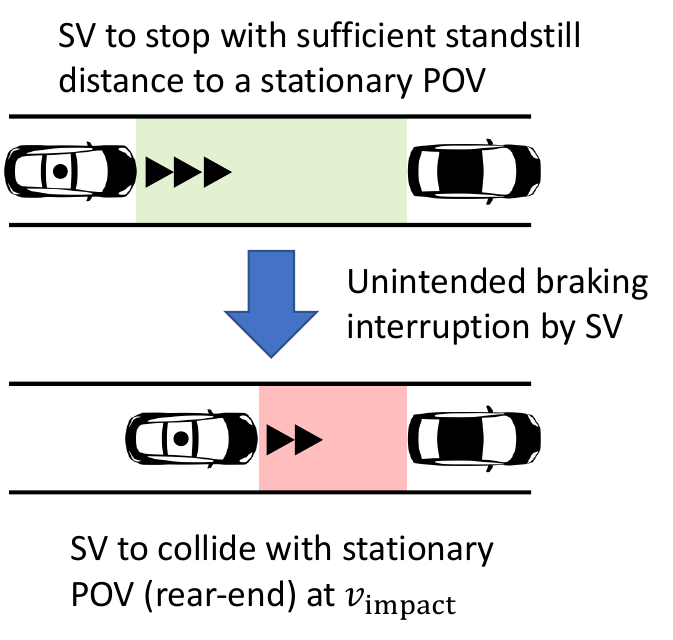}
    \caption{The \gls{SV} (left) experiences an unintended braking interruption (\gls{UBI}) when braking for a \glsdisp{POV}{POV} (right) stopped ahead. The \gls{UBI} transforms a safe situation (top) into an unsafe one (bottom).}
    \label{fig:braking_scenario}
\end{figure}

We illustrate the key ideas of severity evaluation in \gls{MoSAFE} for the intended behavior of ``braking for a stationary vehicle ahead'', which we refer to as the \glsfirst{POV}, and the \gls{HB} of unintended braking interruption (\gls{UBI}) (see Fig.~\ref{fig:braking_scenario}). The intended behavior for the \gls{SV} is to brake at a comfortable level $a_\text{b,min}$ to stop at a required standstill distance $\Delta{s}_\text{stand}$ behind \gls{POV}. An \gls{HB} such as a \gls{UBI} would cause the \gls{SV} to approach the \gls{POV} too fast, so that a higher level of braking would be necessary to stop at $\Delta{s}_\text{stand}$ behind the \gls{POV}. If the required braking to stop behind the \gls{POV} without colliding with it exceeds the maximum braking capacity $a_\text{max}$ of the \gls{SV}, the \gls{UBI} becomes hazardous and will lead to a rear-end collision at a certain collision velocity $v_\text{impact}$, which then can be mapped to collision severity using the Delta-V method.

\begin{figure}[h]
\centering
\begin{subfigure}{0.4\textwidth}
    \includegraphics[width=\textwidth]{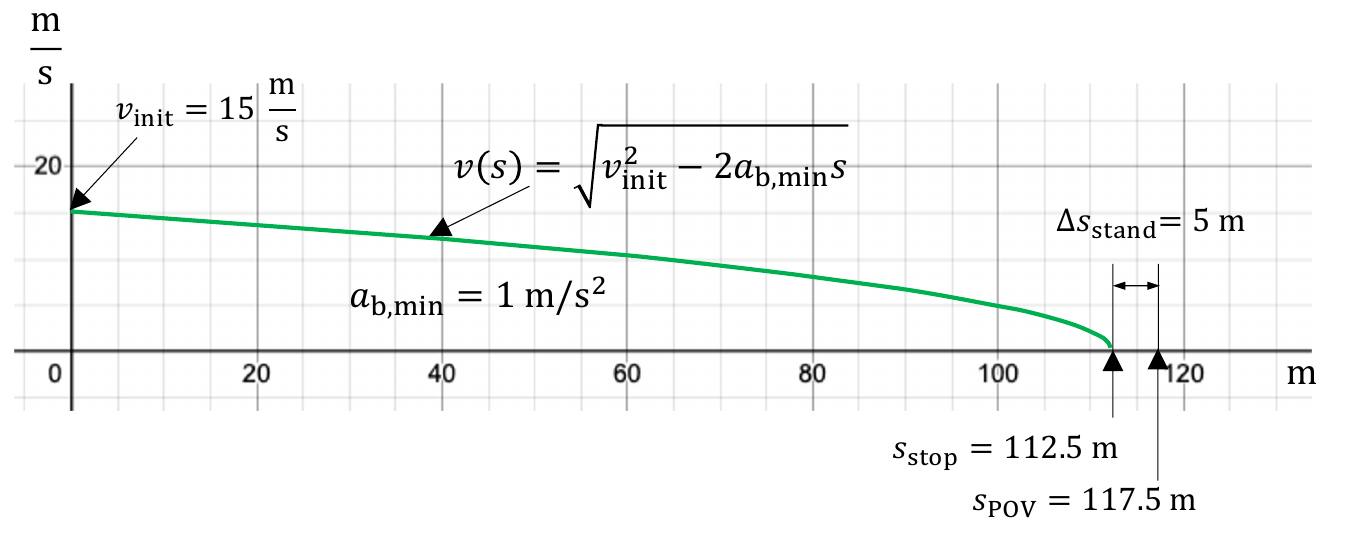}
    \vspace*{-10mm}
    \caption{Nominal braking}
    \vspace*{5mm}
    \label{fig:nominal_braking}
\end{subfigure}
\hfill
\begin{subfigure}{0.4\textwidth}
    \includegraphics[width=\textwidth]{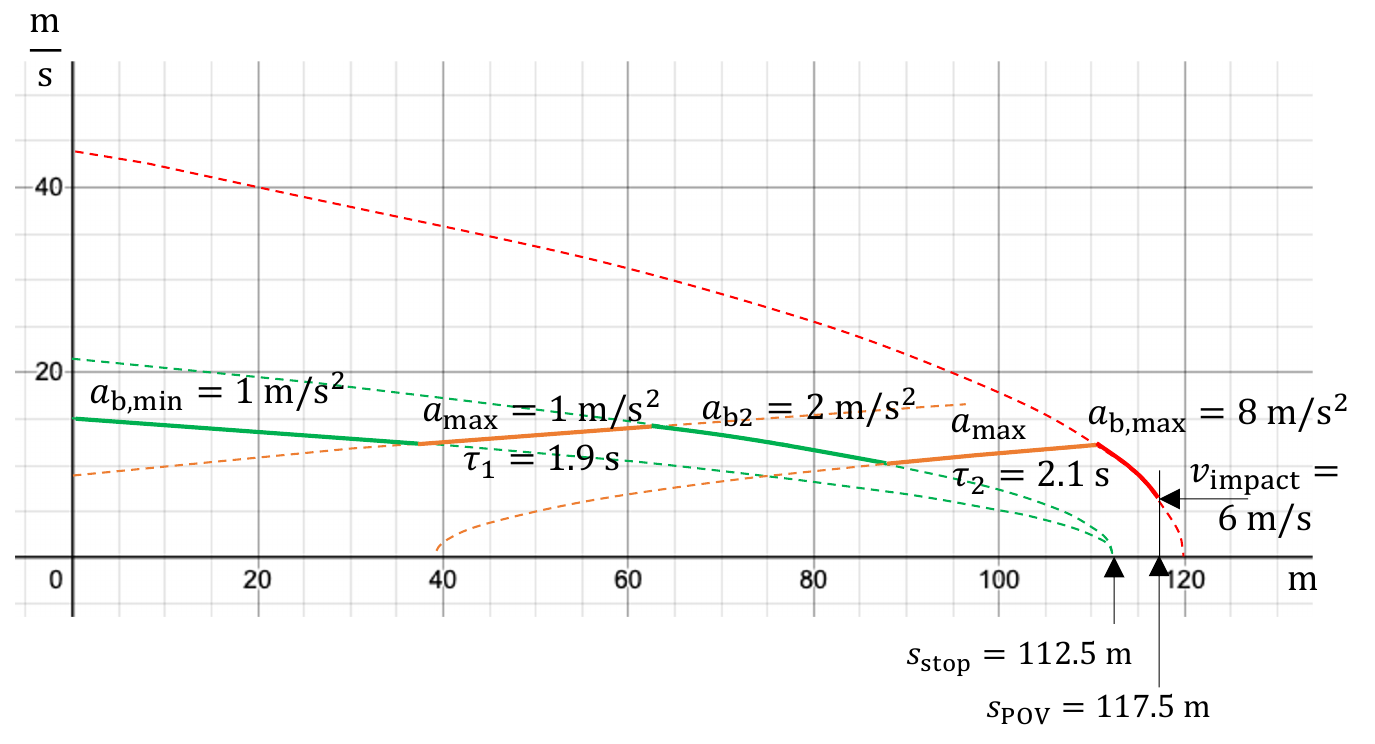}
    \vspace*{-10mm}
    \caption{Multiple \gls{UBI} intervals}
    \vspace*{5mm}
    \label{fig:multi_interrupted_braking}
\end{subfigure}
\hfill
\begin{subfigure}{0.4\textwidth}
    \includegraphics[width=\textwidth]{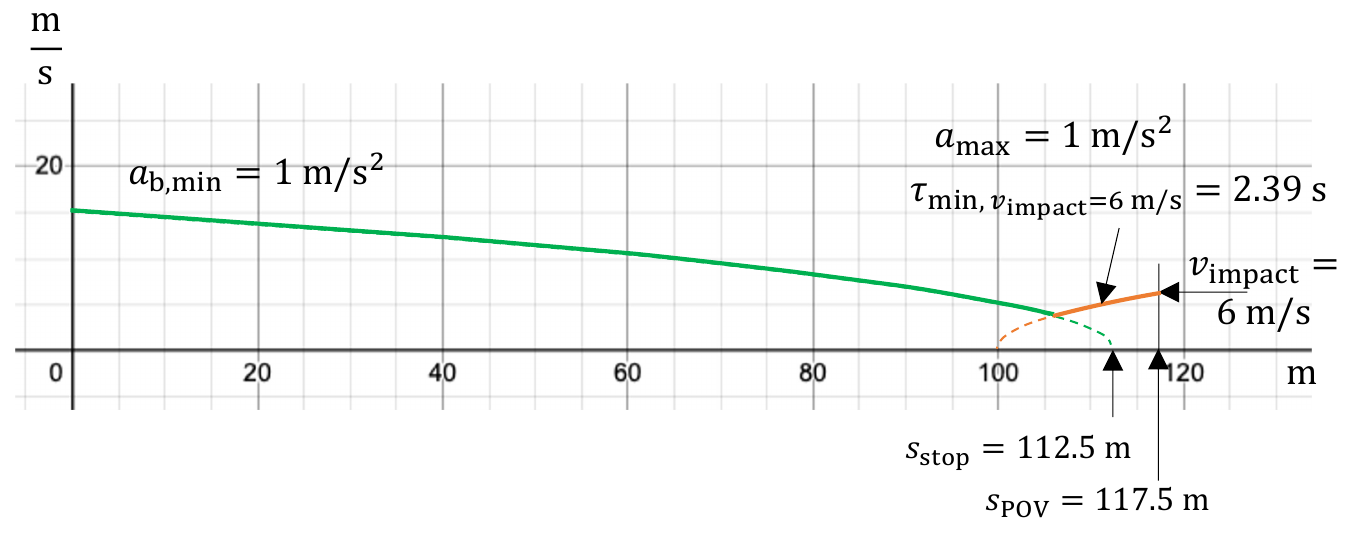}
    \vspace*{-10mm}
    \caption{Shortest \gls{UBI} with $\tau_{\text{min},v_\text{impact}}$ for $v_\text{impact}$}
    \vspace*{5mm}
    \label{fig:optimally_interrupted_braking}
\end{subfigure}
\hfill
\begin{subfigure}{0.4\textwidth}
    \includegraphics[width=\textwidth]{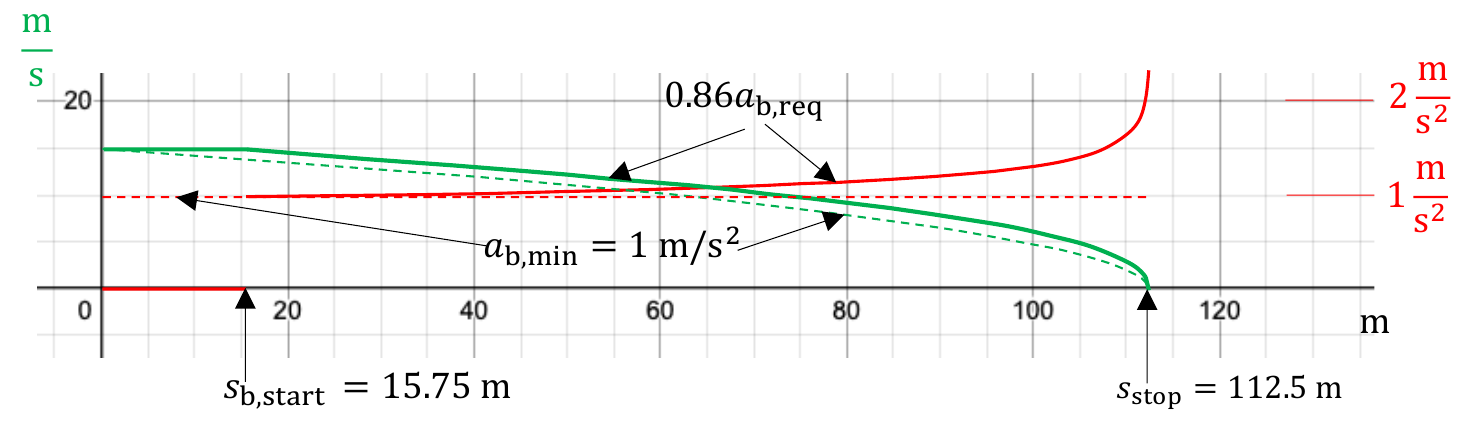}
    \caption{Reduced braking with $0.86a_\text{b,req}$}
    \label{fig:reduced_braking}
\end{subfigure}
        
\caption{Sample braking speed profiles for $v_\text{init}=15\,m/s$ and $s_\text{stop}=112.5\,m$}
\label{fig:profiles}
\end{figure}

\textbf{Establishing high-level scenario models to determine $v_\text{impact}$}. The first step in the severity evaluation is to establish the \gls{HLSM} of the \gls{DAS} and its environment for the given intended behavior (see Fig.~\ref{fig:high_level_model}). The environment of the \gls{DAS} is represented by the \gls{RVE} model, which consists of the relevant road environment elements and the \gls{SV} dynamics. It takes the control input from the \gls{DAS} and provides the \gls{DAS} with observations $y$. The \gls{RVE} model captures the \glspl{HBSC} for our example, including the presence of the stationary \gls{POV}, modeled by its position $s_\text{POV}$; and the \gls{SV} kinematics, modeled by a double integrator $\frac{1}{s^2}$, with the \gls{SV}'s acceleration $a$ as control input and its speed $\dot{s}$ and position $s$ as output. The initial position of \gls{SV} is $s_\text{init}=0$, and its initial speed is $v_\text{init}=v_\text{max}$, to allow for a full braking scenario from $v_\text{max}$ of the \gls{DAS}. The \gls{POV} position is such that the \gls{SV} needs to brake with constant $a_\text{b,min}$ to stop at $s_\text{stop}=s_\text{POV}-\Delta{s}_\text{stand}$ (see Fig.~\ref{fig:nominal_braking}); thus, we have:
\begin{equation}
    s_\text{POV}=s_\text{stop}+\Delta{s}_\text{stand}
    ~~~~~
    s_\text{stop}=\frac{v_\text{init}^2}{2a_\text{b,min}}
\end{equation}

The output $y$ of the \gls{RVE} model is the \gls{SV}'s speed $\dot{s}$ and its distance $d=s_\text{POV}-s$ to the \gls{POV}. The following time-continuous linear \gls{ODE} gives the state-space representation of the model, where $x$ is the system state and $x^{[0]}$ is the initial state at $t=0$: 
\begin{equation}
    \begin{aligned}
        &x=\begin{bmatrix}s\\ \dot{s}\end{bmatrix}~~~~~~y=\begin{bmatrix}d\\ \dot{s}\end{bmatrix}\\
        &\dot{x}=\begin{bmatrix}0&1\\0&0\end{bmatrix}x+\begin{bmatrix}0\\1\end{bmatrix}a\\
        &y=\begin{bmatrix}-1&0\\0&1\end{bmatrix}x+\begin{bmatrix}s_\text{POV}\\0\end{bmatrix}\\
        &x^{[0]}=\begin{bmatrix}0\\v_\text{init}\end{bmatrix}
    \end{aligned}
    \label{eq:environment}
\end{equation}

The intended behavior of the \gls{SV} for our example is represented by a driving policy $\pi$ that applies the required braking $a_\text{b,req}$ to stop at $\Delta{s}_\text{stand}$ behind the \gls{POV}, when $a_\text{b,req}$ is between $a_\text{b,min}$ and $a_\text{b,max}$, where
\begin{equation}
a_\text{b,req}=\frac{\dot{s}^2}{2(d-\Delta{s}_\text{stand})}
\label{eq:areq}
\end{equation}
When $a_\text{b,req}$ is less than $a_\text{b,min}$, the \gls{SV} is free to accelerate to $v_\text{max}$ and continue at that speed. If $a_\text{b,req}$ reaches or exceeds $a_\text{b,max}$, the \gls{SV} will apply $a_\text{b,max}$. Also, the \gls{SV} will apply $a_\text{b,max}$ whenever it moves closer to the \gls{POV} than $\Delta{s}_\text{stand}$. These cases are captured by the following policy function:
\begin{equation}
    \begin{aligned}
        &a=\pi(d,\dot{s})=\\
        &\begin{cases}
      0, &\text{if}~d>\Delta{s}_\text{stand}\land a_\text{b,req}<a_\text{b,min}\land \dot{s}\geq v_\text{max}\\
      a_\text{max}, &\text{if}~d>\Delta{s}_\text{stand}\land a_\text{b,req}<a_\text{b,min}\land \dot{s}<v_\text{max}\\
      -a_\text{b,req}, &\text{if}~d>\Delta{s}_\text{stand}\land a_\text{b,min}\leq a_\text{b,req}<a_\text{b,max}\\
      -a_\text{b,max}, &\text{if}~d>\Delta{s}_\text{stand}\land a_\text{b,req}>a_\text{b,max}\\
      0, &\text{if}~d\leq\Delta{s}_\text{stand}\land \dot{s}=0\\
      -a_\text{b,max}, &\text{if}~d\leq\Delta{s}_\text{stand}\land \dot{s}>0
        \end{cases}
    \end{aligned}
    \label{eq:policy}
\end{equation}
Note that this policy has discrete transitions in acceleration to simplify the analysis, but these transitions would lead to jerky driving. An actually implemented policy would have smooth transitions, but its braking level would need to be close to $a_\text{b,req}$ on average, and therefore the discrete policy is adequate for evaluating the safety of applying $a_\text{b,req}$ as an intended target and the safety of deviating from it. The adequacy of analyzing a smooth policy using a discrete approximation is confirmed by the model validation results using simulation testing on p.~\pageref{sec:sim}. 

\textbf{Evaluation of the intended behavior}. The intended behavior represented by this policy under the modeled \glspl{HBSC} is safe, as the vehicle is guaranteed to stop at $x_\text{stop}$, i.e., $\Delta{s}_\text{stand}$ behind the \gls{POV} (Fig.~\ref{fig:nominal_braking}). The combination of the \gls{RVE} model (\ref{eq:environment}) and the policy (\ref{eq:policy}) (i.e., the \gls{HLSM} in Fig.~\ref{fig:high_level_model} where the red part is ignored) results in the following non-linear \gls{ODE} during braking:
\begin{equation}
\Ddot{s}=\frac{\dot{s}^2}{2(s_\text{stop}-s)}
\label{eq:ODE}
\end{equation}

Assuming the initial condition $s^{[0]}=0$ and $\dot{s}^{[0]}=v_\text{init}$, this \gls{ODE} can be shown to have a solution that corresponds to the application of constant acceleration $a_\text{b,min}$, resulting in the speed profile in Fig.~\ref{fig:nominal_braking}.

Figure~\ref{fig:high_level_model} also includes a crash check $\phi(y)$, which has two components: $\phi_1(y)$ checks whether the \gls{SV} has collided with the \gls{POV}, and $\phi_2(y)$ keeps track of $v_\text{impact}$:
\begin{equation}
    \begin{aligned}
    &\phi_1(d,\dot{s})=I_{\exists t\in[0..T]:d^{[t]}\leq 0}\\
    &v_\text{impact}=\phi_2(d,\dot{s})=\dot{s}^{[t_c]},t_c=\text{min}~t\in[0..T], d^{[t]}\leq 0
    \end{aligned}
\end{equation}
where $I$ is the indicator function, and $T$ is duration of the braking scenario. Note that safety properties such as $\phi_1(y)$ could be expressed using a temporal logic, such as Signal Temporal Logic (STL)~\cite{Maler04}, but we choose to use first-order logic for simplicity.

\textbf{Evaluation of the \glspl{HB}}. Having evaluated the intended behavior as safe, we turn to evaluating the severity of the \glspl{HB}, which are hazardous deviations for the intended behavior. This is achieved by injecting the \glspl{HB} into the intended behavior of the \gls{HLSM}. For \gls{UBI}, the injection is accomplished by adding the switch $\iota$ in Fig.~\ref{fig:high_level_model}, which interrupts braking by injecting $a_\text{max}$ when the \gls{SV}'s speed is below $v_\text{max}$ or zero acceleration otherwise:
\begin{equation}
    \text{max\_acc}(\dot{s})=
    \begin{cases}
        a_\text{max}, &\text{if}~\dot{s}<v_\text{max}\\
        0, &\text{if}~\dot{s}\geq v_\text{max}
    \end{cases}
\end{equation}
The switch $\iota$ operates in discrete time and is controlled by the sequence $\rho_a\subseteq\mathbb{N}$, which contains the time steps for which the switch should be in on-position, i.e., connecting to $h$ rather than $h'$, and injecting a braking interruption. The current time step $k$ is computed as the integer part, represented by the floor operator, of the current continuous time $t$ divided by the time-step duration $\Delta{t}$. Formally, the switch is defined by the following function:
\begin{equation}
    \begin{aligned}
        &\iota^{[t]}(h,h',\rho_a)=
        \begin{cases}
            h, &\text{if}~k(t)\in\rho_a\\
            h', &\text{if}~k(t)\notin\rho_a
        \end{cases}
    \end{aligned}~~~~
    \begin{aligned}
        &\rho_a\subseteq\mathbb{N}\\
        &k(t)=\floor*{\frac{t}{\Delta{t}}}
    \end{aligned}
\end{equation}
The switch allows injecting an arbitrary sequence of braking interruptions, up to the time resolution $\Delta{t}$, which can be set as finely as needed. Figure~\ref{fig:multi_interrupted_braking} shows an example speed profile resulting from injecting $\rho_a=\{26..45, 66..87\}$ with $\Delta t=$0.1\,s. This sequence injects two braking interruption intervals, the first one with the duration $\tau_1=$1.9\,s and the second one with $\tau_2=$2.1\,s. The first interruption changes the approach situation of the \gls{SV}, requiring it to apply $a_\text{b,req}=$2\,m/s$^2$, rather than the initial $a_\text{b,min}=$1\,m/s$^2$. The second interruption puts the \gls{SV} on a maximum braking trajectory with $a_\text{b,max}=$8\,m/s$^2$ to crash into the \gls{POV} with $v_\text{impact}=$6\,m/s.    

Evaluating the severity of a multi-interval braking interruption is complex, as the number of the intervals, their start times, and their duration influence the resulting $v_\text{impact}$. This complexity likely arises for many other \glspl{HB}, where a multitude of specific \gls{HB} sequences need to be mapped to the resulting $v_\text{impact}$.

A general solution strategy in \gls{MoSAFE} is to employ a conservative over-approximation, where a class of sequences $\mathbb{P}\subseteq2^\mathbb{N}$, specified by an abstract pattern, is bounded by the maximum $v_\text{impact}$ that any of the sequences in $\mathbb{P}$ can result in. For \gls{UBI}, we define the pattern $\mathbb{P}_{a,k_\text{min},k_\text{max}}$ to denote all sequences $\rho_a$ with the total duration of braking interruption being at least $k_\text{min}$ and at most $k_\text{max}$ time steps. The lower bound helps eliminate \gls{UBI} sequences that are guaranteed to be safe. Without any loss of generality, these sequences are also limited by the maximum duration $T_\text{max}$ of an approach scenario. This maximum occurs in the nominal case of braking with $a_\text{b,min}$ (Fig.~\ref{fig:nominal_braking}), since any \gls{UBI} would lead to higher approach velocities and thus shorter approach time:
\begin{equation}
T_\text{max}=\frac{v_\text{init}}{ a_\text{min}}
\end{equation}
We also define the corresponding maximum duration $n_\text{max}$ in discrete steps (the ceiling operator ensures that $n_\text{max}$ covers $T_\text{max}$ completely):
\begin{equation}
n_\text{max}=\ceil*{\frac{T_\text{max}}{\Delta t}}
\end{equation}
The \gls{UBI} pattern $\mathbb{P}_{a,\tau_\text{min},\tau_\text{max}}$ is formally specified as follows:
\begin{equation}
    \mathbb{P}_{a,k_\text{min},k_\text{max}}=\{\rho_a\subseteq\mathbb{N}_{n_\text{max}}|k_\text{min}\leq|\rho_a|\leq k_\text{max}\}
    \label{eq:UBI-pattern}
\end{equation}
where $\mathbb{N}_n=\{k\in\mathbb{N}|k\leq n\}$ and $|\rho_a|$ denotes the size of $\rho_a$ and corresponds to the number of discrete time-steps where \gls{UBI} is injected.

Given an impact velocity $v_\text{impact}$, there exists the shortest total duration $\tau_{\text{min},v_\text{impact}}$ of \gls{UBI} that results in a crash with $v_\text{impact}$, and this duration can be computed in closed form. The details of this statement are beyond the scope of this paper, but it relies on a theorem that multiple \gls{UBI} intervals can always be replaced by a single one that results in a higher $v_\text{impact}$ than the multiple ones. Thus, $\tau_{\text{min},v_\text{impact}}$ can be computed by minimizing the duration of a single \gls{UBI} interval for a given $v_\text{impact}$. For example, Fig.~\ref{fig:optimally_interrupted_braking} shows the shortest \gls{UBI} interval $\tau_{\text{min},v_{\text{impact}}=6\,\text{m/s}}=$2.39\,s that results in the same $v_{\text{impact}}=$6\,m/s as the multiple \gls{UBI} intervals with $\tau_\text{total}=\tau_1+\tau_2=$4\,s in Fig.~\ref{fig:multi_interrupted_braking}.

\begin{table}[h]
    \centering
    \begin{tabular}{|p{0.03\textwidth}|p{0.18\textwidth}|p{0.20\textwidth}|} \hline 
         \textbf{Seve\-rity class}& $v_\text{impact}$ \textbf{range (m/s)} & $\tau_\text{total}$ \textbf{range (s)}\\ \hline 
         S0& $[v_\text{min}..v_\text{S0}]=[0..5.3]$ & $[\tau_\text{contact}..\tau_\text{S0}]=[1.97..2.29]$ \\ \hline 
         S1& $(v_\text{S0}..v_\text{S1}]=(5.3..7.8]$ & $(\tau_\text{S0}..\tau_\text{S1}]=(2.29..2.76]$\\ \hline 
         S2& $(v_\text{S1}..v_\text{S2}]=(7.8..10.3]$ & $(\tau_\text{S1}..\tau_\text{S2}]=(2.76..3.57]$\\ \hline 
         S3& $(v_\text{S2}..v_\text{max}]=(10.3..15]$ & $(\tau_\text{S2}..\tau_\text{max}]=(3.57..7.83]$\\ \hline
    \end{tabular}
    \caption{Range of \gls{UBI} duration $\tau_\text{total}$ that could result in a crash with a given maximum severity, assuming $a_\text{b,min}$=1\,m/$\text{s}^2$, $a_\text{b,max}$=8\,m/$\text{s}^2$, $a_\text{max}$=1\,m/$\text{s}^2$,
    $v_\text{init}$=15\,m/s,
    $s_\text{POV}$=117.5\,m, and
    $\Delta{s}_{\text{stand}}$=5\,m } 
    \label{tab:severity}
\end{table}

\textbf{Mapping collision configuration and $v_\text{impact}$ to severity classes}. The next step is to determine the range of $v_\text{impact}$ for each of the S0-3 severity classes, which will allow us to determine the maximum severity class for a given total duration $\tau_\text{total}$ of \gls{UBI} (Tab.~\ref{tab:severity}). 
The \gls{HLSM} gives us the crash configuration and $v_\text{impact}$ due to the injected \gls{HB}. For \gls{UBI}, the configuration is a front-to-rear collision, and $\tau_\text{total}$ puts an upper bound on $v_\text{impact}$. The crash configuration and $v_\text{impact}$ allow us to estimate the severity class based on statistical injury models according to the guidance in SAE J2980~\cite{J2980}. As an example, Tab.~\ref{tab:severity} gives the severity class (first column) based on injury risk to belted front-row occupants of the \gls{SV} during a front-to-rear collision at $v_\text{impact}$ that falls into the ranges specified in the second column. For example, when $v_\text{impact}$ falls into the range $[v_\text{min}..v_\text{S0}]=[0..5.3]$\,m/s the corresponding injury is estimated as severity class S0, where $v_\text{min}=0$ is the minimum $v_\text{impact}$ (when the \gls{SV} just touches the \gls{POV}) and $v_\text{S0}=$5.3\,m/s is the maximum $v_\text{impact}$ that still results in S0. 
Table~\ref{tab:severity} (in this paper) is based on an existing statistical model~\cite{krampe20injury}; in particular, the severity classes are defined in Table 2 there \cite{krampe20injury}, and the delta V values for each corresponding traffic domain are specified in Table 3 there \cite{krampe20injury}. Conservatively, delta V, which is the change of velocity of the bullet vehicle due to the collision, is equated with $v_\text{impact}$; in reality, delta V is about half of $v_\text{impact}$ if \gls{SV} and \gls{POV} have the same mass, and delta V approaches $v_\text{impact}$ when \gls{SV} strikes a heavy vehicle, such as a truck or a bus~\cite{bonnett2001stiffness}. Table~\ref{tab:severity} is also consistent with another model of severe injury~\cite{jurewicz16impact}, which estimates probability of \glsxtrfull{MAIS} 3+ injury as a function of $v_\text{impact}$ (see Fig.~4 in ~\cite{jurewicz16impact}). For example, S2 corresponds to more than 10\% of \gls{MAIS} 3+ (as defined in Table B.1 in \cite{ISO26262}, Part 3), and the model~\cite{jurewicz16impact} predicts this risk at a critical impact speed of about 30\,km/h (8.3\,m/s). 

Given the ranges of $v_\text{impact}$ for each severity class, we can map them to the corresponding ranges of the shortest total duration $\tau_{\text{min},v_\text{impact}}$ (third column in Tab.~\ref{tab:severity}). For example, $[v_\text{min}..v_\text{S0}]$ in the first row is mapped to $[\tau_\text{contact}..\tau_\text{S0}]$, where $\tau_\text{contact}=\tau_{\text{min},v_\text{impact}=0}$ represents the shortest duration of \gls{UBI} for a crash at $v_\text{impact}=0$, and $\tau_\text{S0}=\tau_{\text{min},v_\text{impact}=v_\text{S0}}$ represents the shortest duration of \gls{UBI} for a crash that still results in S0. As preciously described, the shortest duration $\tau_{\text{min},v_\text{impact}}$ for a given $v_\text{impact}$ can be computed as an optimization solution in closed form. As another example, the interval $(v_\text{S2}..v_\text{max}]$ of impact velocities that would likely result in S3 is mapped to $(\tau_\text{S2}..\tau_\text{max}]$, where $\tau_\text{S2}=\tau_{\text{min},v_\text{impact}=v_\text{S2}}$ represents the shortest duration of \gls{UBI} for a crash that may result in a crash with maximum severity of S2, and $\tau_\text{max}$ is the maximum duration of \gls{UBI}, which occurs when the \gls{SV} continues at $v_\text{max}$ without braking, i.e., $\tau_\text{max}=s_\text{POV}/v_\text{max}\approx$ 7.83\,s. Note that a bracket marks an interval bound that is included in the interval, and a parenthesis marks a bound that is excluded.

\begin{table}[ht]
    \centering
    \begin{tabular}{|p{0.06\textwidth}|p{0.16\textwidth}|p{0.17\textwidth}|} \hline 
         \textbf{\gls{UBI} pattern}& \textbf{Definition} & \textbf{Description}\\ \hline 
         $\mathbb{P}_{a,\text{nocrash}}$& $\mathbb{P}_{a,k_\text{min}=0,k_\text{max}=k_\text{contact}-1}$ & All \gls{UBI} sequences guaranteeing no crash \\ \hline 
         $\mathbb{P}_{a,\text{S0..3}}$& $\mathbb{P}_{a,k_\text{min}=k_\text{contact},k_\text{max}=n_\text{max}}$ & All \gls{UBI} sequences that may lead to a crash\\ \hline 
         $\mathbb{P}_{a,\text{S1..3}}$& $\mathbb{P}_{a,k_\text{min}=k_\text{S0}+1,k_\text{max}=n_\text{max}}$ & All \gls{UBI} sequences that may lead to a crash at severity S1 or higher\\ \hline 
         $\mathbb{P}_{a,\text{S2..3}}$& $\mathbb{P}_{a,k_\text{min}=k_\text{S1}+1,k_\text{max}=n_\text{max}}$ & All \gls{UBI} sequences that may lead to a crash at severity S2 or higher\\ \hline
         $\mathbb{P}_{a,\text{S3}}$& $\mathbb{P}_{a,k_\text{min}=k_\text{S2}+1,k_\text{max}=n_\text{max}}$ & All \gls{UBI} sequences that may lead to a crash at severity S3\\ \hline         
    \end{tabular}
    \caption{\gls{UBI} patterns by severity crash they can cause; note that $k_\text{contact}=k(\tau_\text{contact})$, $k_\text{S0}=k(\tau_\text{S0})$, $k_\text{S1}=k(\tau_\text{S1})$, and $k_\text{S2}=k(\tau_\text{S2})$} 
    \label{tab:UBI}
\end{table}

\begin{figure*}[ht]
    \centering
    \includegraphics[width=0.8\textwidth]{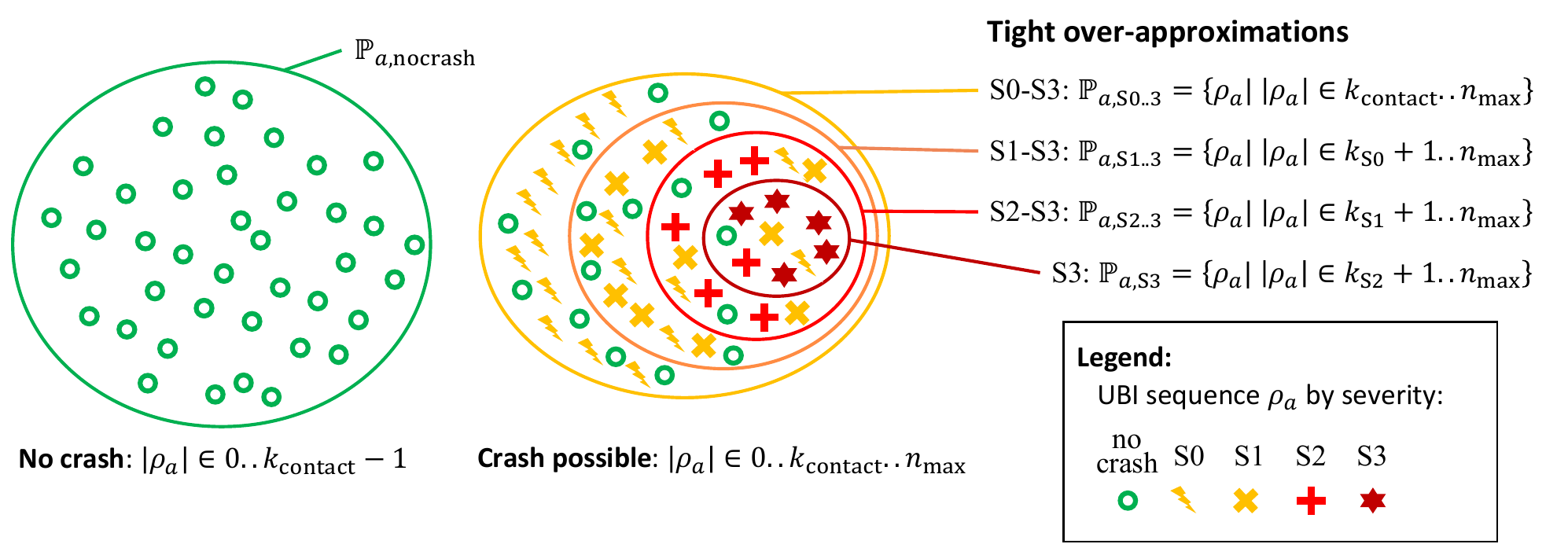}
    \caption{Illustration of the \gls{UBI} patterns from Tab.~\ref{tab:UBI} as sets of \gls{UBI} sequences. Each glyph in the Venn diagram represents a particular \gls{UBI} sequence $\rho_a$ and its color indicates the severity of a crash that it would cause in the analyzed scenario. Note that the hazardous patterns are tight over-approximations of their corresponding severity range; e.g., $\mathbb{P}_{a,\text{S3}}$ contains all S3 sequences, but it also contains sequences of all other severities. The patterns are tight given their simple form as in eq.~\ref{eq:UBI-pattern}.}
    \label{fig:overapproximation}
\end{figure*}

\textbf{Specification of \glspl{HBP}.} The $\tau_\text{total}$ ranges in Tab.~\ref{tab:severity} allow defining \gls{UBI} patterns that are bounded by severity (see Tab.~\ref{tab:UBI}). The first pattern in Tab.~\ref{tab:UBI}, $\mathbb{P}_{a,\text{nocrash}}$, contains all \gls{UBI} sequences guaranteeing no crash, i.e., those with $\tau_\text{total}$ shorter than $\tau_\text{contact}$. Thus, this \gls{UBI} pattern summarizes all \gls{UBI} sequences that are non-hazardous. The remaining patterns may lead to crashes and thus are hazardous, i.e., they are \glspl{HBP} (Fig.~\ref{fig:steam}). The first \gls{HBP} (second row in Tab.~\ref{tab:severity}), $\mathbb{P}_{a,\text{S0..3}}$, contains all hazardous \gls{UBI} sequences, i.e., all \gls{UBI} sequences that may lead to a crash. The next \gls{HBP}, $\mathbb{P}_{a,\text{S1..3}}$, contains all \gls{UBI} sequences that may lead to a crash at severity S1 or higher. The remaining two \glspl{HBP} are defined analogously. Since we consider the probability occurrence of the pattern during the braking scenario and the scenario duration varies depending on the number of \gls{UBI} time steps, we use the longest duration $T_\text{max}$ and set $k_\text{max}$ to $n_\text{max}$, instead of $k(\tau_\text{max})$.

Note that each of the hazardous \gls{UBI} patterns in Tab.~\ref{tab:UBI} contains all of the \gls{UBI} sequences that lead to a crash within the pattern's severity range, but may also include \gls{UBI} sequences of a lesser severity (see~Fig.~\ref{fig:overapproximation}). In fact, they are the tightest over-approximations wrt. their specified severity range, given the pattern form in eq.~\ref{eq:UBI-pattern}. For example, $\mathbb{P}_{a,\text{S2..3}}$ contains all \gls{UBI} sequences that lead to a crash at severity S2 or higher and so does $\mathbb{P}_{a,\text{S1..3}}$. Further, while $\mathbb{P}_{a,\text{S1..3}}$ also includes all \gls{UBI} sequences that lead to crashes at severity S1, some of these may also be in $\mathbb{P}_{a,\text{S2..3}}$, and both patterns may include \gls{UBI} sequences of severity S0 or even non-hazardous ones. However, $\mathbb{P}_{a,\text{S2..3}}$ is a tighter over-approximation of all \gls{UBI} sequences that lead to a crash at severity S2 or higher than $\mathbb{P}_{a,\text{S1..3}}$. 

In our sample scenario (Fig.~\ref{fig:nominal_braking}), assuming $\Delta t=$0.1\,s, injecting $\mathbb{P}_{a,\text{S0..3}}$ corresponds to the occurrence of $k_\text{min}$=19 or more time steps with \gls{UBI} within $n_\text{max}$=150 time steps of the maximum scenario duration. This is because the duration $T_\text{max}$ of the nominal scenario in Fig.~\ref{fig:nominal_braking} is 15\,s, i.e., $n_\text{max}$=150, and $k_\text{min}=k_\text{contact}=k(1.97\,\text{s})=19$ (from the first row in Tab.~\ref{tab:severity} and the second row in Tab.~\ref{tab:UBI}). Thus, if \gls{UBI} occurs in fewer than 19 out of 150 time steps, no collision will occur due to \gls{UBI}. Similarly, $\mathbb{P}_{a,\text{S3}}$ tells us that no collision of severity S3 can occur in our scenario due to \gls{UBI} if \gls{UBI} occurs in fewer than $k_\text{S2}$=35+1=36 out of 150 time steps.

Limiting the likelihood of occurrence of the hazardous \gls{UBI} patterns in Tab.~\ref{tab:UBI} allows limiting the safety risk of crashes due to \gls{UBI}. These likelihood limits, after being multiplied by exposure, would be assigned based on acceptance criteria. 

\subsection{Identification and Evaluation of Functional Insufficiencies \& Triggering Conditions (Clause 7)}
\label{subsec:clause7}

Clause 7 focuses on the identification and evaluation of functional insufficiencies and triggering conditions (the left side of Fig.~\ref{fig:steam}). Functional insufficiencies include (i) insufficiencies of specification and (ii) performance insufficiencies at the vehicle level and element level. For our example, functional insufficiencies at the vehicle level have already been addressed in the previous section: (i) the specification of the driving policy for our sample intended behavior has been shown to be safe and (ii) the target performance at the vehicle level is given by the acceptance criteria that assign an upper bound on the probability of a collision due to \gls{UBI}. That probability would be further decomposed into upper bounds on the probability of \gls{UBI} occurrence in different \glspl{HBSC} (see \cite{ISCaP} for more detail on this decomposition). Thus, this section will focus on the functional insufficiencies at the element level and their triggering conditions.

\subsubsection{Model-based Identification and Evaluation of Functional Insufficiencies at the Element Level}
\label{subsubsec:heps}

The \gls{MoSAFE} method helps identify and evaluate functional insufficiencies using a \gls{DSM}, which is a refinement of the \gls{HLSM} from the previous activity. The \gls{DSM} is used to (i) evaluate the intended behavior of the \gls{DAS} components in the given scenario and (ii) identify, specify, and evaluate hazardous deviations from the intended behavior at the element level as \glspl{HEP}.

\begin{figure}[h]
    \centering
    \includegraphics[width=0.5\textwidth]{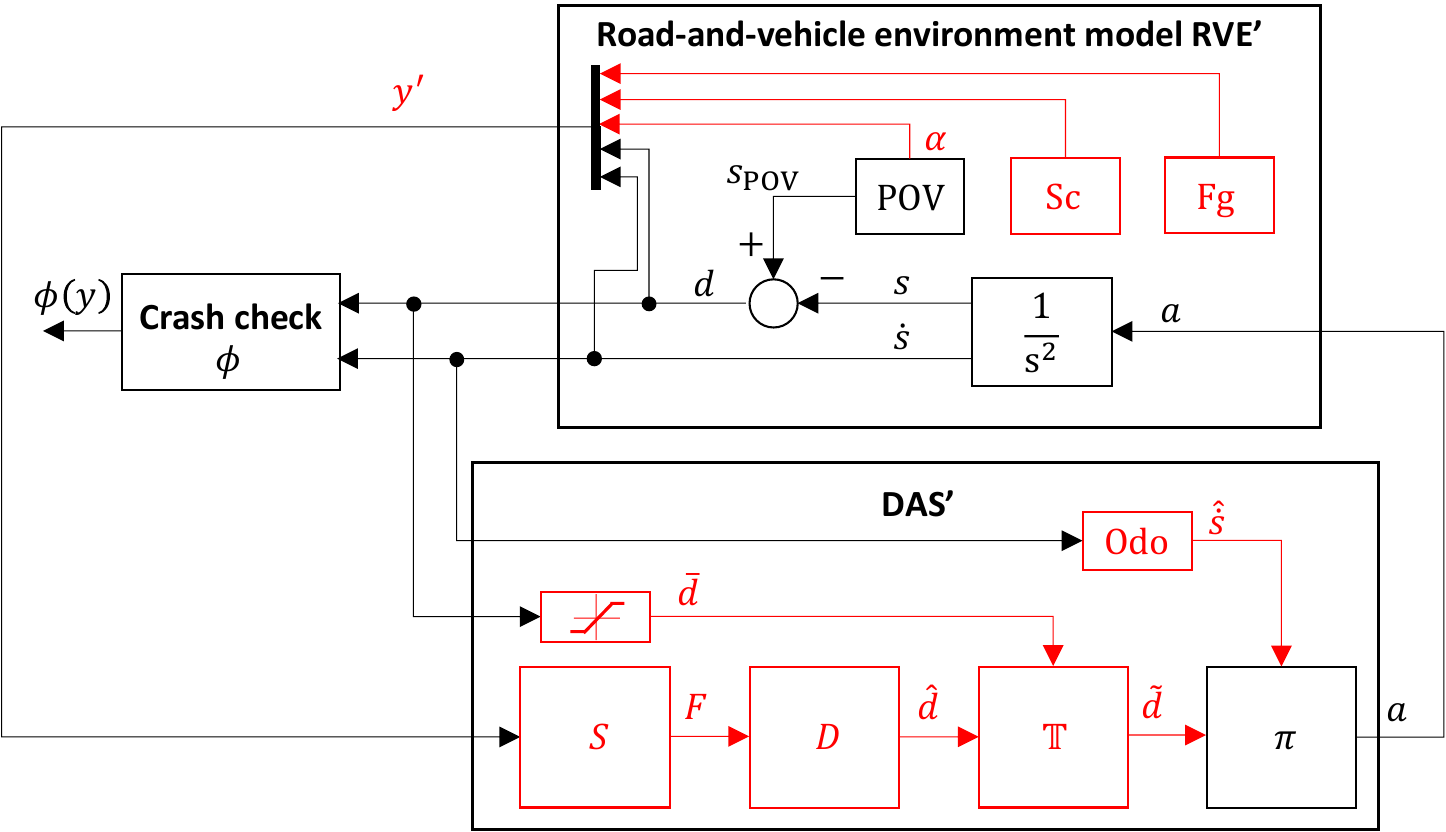}
    \caption{Detailed scenario model for ``braking for a stationary vehicle ahead'', including the refined models RVE' and DAS' (refinements in red)}
    \label{fig:refined_model}
\end{figure}

\textbf{Establishing detailed scenario models}. The first step is to establish the detailed scenario model (\gls{DSM}) for the intended behavior of the \gls{SV} (see Fig.~\ref{fig:refined_model}. To this end, the \gls{RVE} model is refined to include all \glsfirstplural{IRC} (see quadrants 1 and 2 in Fig.~\ref{fig:steam}), not just those \glspl{IRC} that are also \glspl{HBSC} (quadrant 2) and thus already included in the \gls{HLSM}. In our example, the kinematic model from Fig.~\ref{fig:high_level_model} is augmented with three sample \glspl{IRC} being the \gls{POV}'s appearance $\alpha$, the remaining scenery \textit{Sc} (such as the road appearance), and fog \textit{Fg}. These \glspl{IRC} may be represented at different levels of abstraction, such as using high-level attributes (typically based on an \gls{ODD} ontology, e.g.,~\cite{ISO34503}) or detailed shape and appearance information. The \gls{DAS} model is refined based on the system design, and it is denoted as DAS'. In our example, the model includes the original driving policy $\pi$ (eq.~\ref{eq:policy}) as the intended behavior of the planning components, and four additional perception components: a sensor $S$, such as a camera or lidar, an object detector $D$, a tracker $\mathbb{T}$, and an odometry component \textit{Odo}. For simplicity, we do not further decompose $\pi$ into planning, control, and actuation elements. All outputs from the refined \gls{RVE} model RVE' are bundled in $y'$ and input into the sensor $S$. They are all relevant to the sensor, potentially affecting its output frame $F$, and are thus \glspl{IRC}. The output frame $F$, such as a camera image or lidar scan, is fed into the object detector $D$, which normally relies on deep neural networks to produce detections. For the purpose of the scenario, the intended behavior of the composition of $S$ and $D$ is to estimate the distance $d$ to the \gls{POV}, limited by the maximum detection range $r_\text{max}$ of the sensor-detector combination:
\begin{equation}
    \hat{{d}}=D(S(y'))\approx\overline{d}, ~~~ 0\leq \hat{{d}}\leq r_\text{max}
\end{equation}
where $\hat{{d}}$ is an estimate of the the ground-truth range-limited distance $\overline{d}$, defined as
\begin{equation}
    \overline{d}=
    \begin{cases}
        d , &\text{if}~0\leq d< r_\text{max} \\
        r_\text{max}, &\text{otherwise}    
    \end{cases}
    \label{eq:range_limited_d}
\end{equation}
Under these definitions, $\hat{{d}}<r_\text{max}$ means that an object is detected, and $\hat{{d}}=r_\text{max}$ means that no object is detected. If $\hat{{d}}=r_\text{max}$ even though $\overline{d}<r_\text{max}$, we have a case of a \glsfirst{FN} detection. Conversely, if $\hat{{d}}<r_\text{max}$ even though $\overline{d}=r_\text{max}$, we have a \glsfirst{FP} detection. Otherwise, we have a \glsfirst{TP} detection, with the estimate $\hat{{d}}$ corrupted by a distance error $\epsilon_d=\hat{{d}}-\overline{d}$. The sensor-detector combination is complex and has a stochastic nature. Even though faults in the detector model are systematic, its input $F$ is stochastic and conditioned on (i) the sensor characteristics, including random hardware noise, such as camera shot noise, and (ii) the \glspl{IRC}, many of which are specified with uncertainty (because of their high-dimensional nature, such as appearance and weather). Thus, the sensor-detector combination is modeled probabilistically, and its error distribution is estimated through testing:
\begin{equation}
    \hat{{d}}\sim p(\hat{{d}}|y')
\end{equation}
The tracker $\mathbb{T}$ takes a sequence of $c$ latest distance estimates $\hat{{d}}^{[k(t)]},\dots,\hat{{d}}^{[k(t)-c]}$ and produces the best estimate for the current time $t$. While an actual implementation would use use an estimator such as a Kalman filter, we abstract the tracker to focus on its track management logic, which affects how detection errors propagate through the tracker. For our example, the tracker model $\mathbb{T}$ expresses the typical ``keep alive'' logic, where a track is terminated (i.e., returning $r_\text{max}$) when there is no detection for $c$ consecutive time steps (i.e., each of the latest $c$ distance estimates is $r_\text{max}$); otherwise, it returns the ground-truth range-limited distance $\overline{d}$ (for simplicity, our example ignores distance estimate errors):
\begin{equation}
  \begin{aligned}
    \Tilde{{d}}=\mathbb{T}(\hat{{d}}^{[k(t)]},\dots,\hat{{d}}^{[k(t)-c]},\overline{d})=\\
    \begin{cases}
        r_\text{max}, &\text{if}~\hat{{d}}^{[k(t)]}=\dots=\hat{{d}}^{[k(t)-c]}=r_\text{max}\\
        \overline{d}, &\text{otherwise}    
    \end{cases}
  \end{aligned}
  \label{eq:tracker}
\end{equation}
Note that while $\mathbb{T}$ samples $\hat{{d}}$ for $c$ discrete time steps, it outputs $\Tilde{{d}}$ in continuous time, as expected by the \gls{DSM}. Also, the range limiter block in Fig.~\ref{fig:refined_model} implements eq.~\ref{eq:range_limited_d}. Further, we assume conservatively $\hat{{d}}^{[k]}=r_\text{max}$ for $k<0$.

Finally, the odometry \textit{Odo} measures the SV speed. Internally, it consists of a sensor (e.g., a wheel encoder) and an estimator (e.g., a Kalman filter). Its intended behavior is an identity function, but, similarly to object detection, its actual behavior is modeled probabilistically as $p(\hat{\dot{s}}|\dot{s})$.

It is easy to show that the intended behavior of the \gls{DAS} in this refined model is safe. Assuming a sufficient detection range $r_\text{max}>s_\text{POV}$ and the intended behavior of each block as specified, the policy $\pi$ receives the ground-truth $d$ and $\dot{s}$, and thus the resulting \gls{SV} braking behavior is same as for the \gls{HLSM} in Figs.~\ref{fig:high_level_model} and~\ref{fig:nominal_braking}.

\begin{figure*}[h]
    \centering
    \includegraphics[width=1\textwidth]{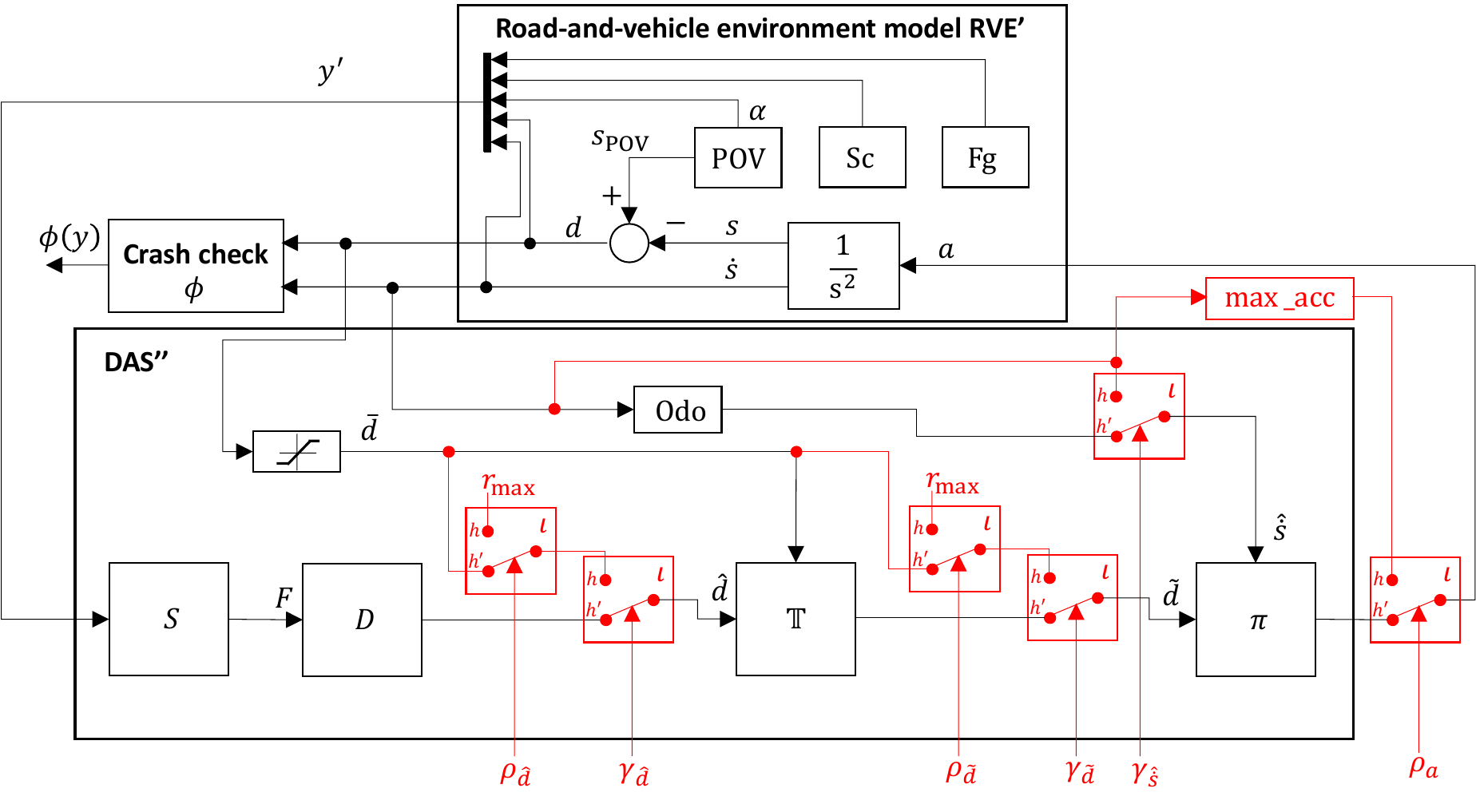}
    \caption{Detailed scenario model (DSM') for ``braking for a stationary vehicle ahead'' with element-level intended behavior and error injection (in red)}
    \label{fig:refined_model_inject}
\end{figure*}

\textbf{Identification and Evaluation of \glspl{HEP}}.
In the next step, the \gls{MoSAFE} method helps to identify, specify, and evaluate hazardous deviations from the intended behavior as \glspl{HEP}. The key idea is to determine error patterns on the input of each component in the DAS' that could lead to a given \gls{HBP} identified in the previous activities (Clause 6). The analysis is performed in a backward direction starting from the \gls{HB} at the output of the DAS' and identifying \glspl{HEP} incrementally component by component towards the inputs $y'$ of the DAS'. In our example (Fig.~\ref{fig:refined_model}), the first step is to determine \glspl{HEP} on the inputs into the policy $\pi$ that would cause the policy to produce a \gls{UBI} pattern of a given severity (Tab.~\ref{tab:UBI}). The next step is to determine \glspl{HEP} on the inputs of the components that feed into the policy, e.g., $\mathbb{T}$, that would cause the \glspl{HEP} on the policy inputs, and so on.

In order to support this analysis, the \gls{DSM} is instrumented to allow injecting intended behavior and error sequences at specific component inputs, resulting in the instrumented DSM' and DAS'' in Fig.~\ref{fig:refined_model_inject}. Intended behavior or error sequences are injected at a specific component input, e.g., $\Tilde{d}$, by inputting specific sequences into the corresponding switches, i.e., $\gamma_{\Tilde{d}}$ and $\rho_{\Tilde{d}}$. For example, no injection at $\Tilde{d}$ occurs when $\gamma_{\Tilde{d}}=\emptyset$; alternatively, the intended behavior for $\Tilde{d}$ is injected when $\gamma_{\Tilde{d}}=\mathbb{N}_{n_\text{max}}\land\rho_{\Tilde{d}}=\emptyset$; and an error sequence from the pattern $\mathbb{P}_{\hat{d}}$ is injected at $\Tilde{d}$ when $\gamma_{\Tilde{d}}=\mathbb{N}_{n_\text{max}}\land\rho_{\Tilde{d}}\in\mathbb{P}_{\hat{d}}$.

\begin{table}[h]
    \centering
    \begin{tabular}{|p{0.05\textwidth}|p{0.15\textwidth}|p{0.19\textwidth}|} \hline 
         \textbf{Name}& \textbf{Definition} & \textbf{Description}\\ \hline 
         $C_\text{DSM}$& $\gamma_{\hat{\dot{s}}}=\gamma_{\Tilde{d}}=\gamma_{\hat{d}}=\rho_a=\emptyset$ & \gls{DSM} (as in Fig.~\ref{fig:refined_model}) \\ \hline 
         $C_{\mathbb{P}_a}$& $\gamma_{\hat{\dot{s}}}=\gamma_{\Tilde{d}}=\mathbb{N}_{n_\text{max}} \land\rho_{\Tilde{d}}=\emptyset\land\rho_a\in\mathbb{P}_a$ & \gls{DSM} with injected $\mathbb{P}_a$ and intended behavior for $\hat{\dot{s}}$ and $\Tilde{d}$ (equivalent to \gls{HLSM} with injected $\mathbb{P}_a$)\\ \hline 
         $C_{\mathbb{P}_{\Tilde{d}}}$ & $\gamma_{\hat{\dot{s}}}=\gamma_{\Tilde{d}}=\mathbb{N}_{n_\text{max}} \land\rho_a=\emptyset\land\rho_{\Tilde{d}}\in\mathbb{P}_{\Tilde{d}}$ & \gls{DSM} with injected $\mathbb{P}_{\Tilde{d}}$ and intended behavior for $\hat{\dot{s}}$ \\ \hline
         $C_{\mathbb{P}_{\hat{d}}}$ & $\gamma_{\hat{\dot{s}}}=\gamma_{\hat{d}}=\mathbb{N}_{n_\text{max}} \land\gamma_{\Tilde{d}}=\rho_a=\emptyset\land\rho_{\hat{d}}\in\mathbb{P}_{\hat{d}}$ & \gls{DSM} with injected $\mathbb{P}_{\hat{d}}$ and intended behavior for $\hat{\dot{s}}$ \\ \hline       
    \end{tabular}
    \caption{Configurations of the instrumented DSM' from Fig.~\ref{fig:refined_model_inject}} 
    \label{tab:configs}
\end{table}

Table~\ref{tab:configs} defines four configurations of the instrumented DSM' that are relevant to the identification of \gls{HEP}. Configuring the instrumented DSM' with the first configuration $C_\text{DSM}$, which turns all injection switches off, results in the original \gls{DSM} from Fig.~\ref{fig:refined_model}. The remaining three configurations allow injecting deviations from the intended behavior specified by patterns. The first of the three, $C_{\mathbb{P}_a}$, allows injecting \gls{UBI} according to $\mathbb{P}_a$, while supplying $\pi$ with ground-truth $\dot{s}$ and $\overline{d}$; this configuration is equivalent to the \gls{HLSM} with \gls{UBI} injection in Fig.~\ref{fig:high_level_model}. The remaining two configurations allow injecting $r_\text{max}$ (see Fig.~\ref{fig:refined_model_inject}), which corresponds to an \gls{FN} error, at the output of the tracker $\Tilde{d}$ and detector $\hat{d}$, respectively. These three configurations help investigate how \glspl{FN} cause \glspl{UBI} and ultimately define \glspl{HEP} for the detector and the tracker as \gls{FN} patterns. 

The first step in the \gls{HEP} identification is to identify a tracker \gls{FN} pattern $\mathbb{P}_{\Tilde{d}}$ of all the \gls{FN} sequences that would cause a particular \gls{UBI} pattern $\mathbb{P}_a$. This can be phrased as determining the \emph{weakest precondition}~\cite{dijkstra75} on a component input to observe a particular output behavior. More precisely, the step is to determine $\mathbb{P}_{\Tilde{d}}$ such that the set of all behaviors of $\Tilde{d}$ under $C_{\mathbb{P}_{\Tilde{d}}}$ is the largest for which the output $a$ of $\pi$ behaves like under $C_{\mathbb{P}_a}$. In other words, we want to find all \gls{FN} sequences $\rho_{\Tilde{d}}\in\mathbb{P}_{\Tilde{d}}$ that if injected at $\Tilde{d}$ using configuration $C_{\mathbb{P}_{\Tilde{d}}}$ would result in the output $a$ of $\pi$ to behave as if \gls{UBI} sequences $\mathbb{P}_a$ were injected at $a$ using $C_{\mathbb{P}_a}$. Formally, we will denote these sequences by the \gls{WPP} on $\Tilde{d}$ for $\mathbb{P}_a$, defined as follows:
\begin{equation}
    \text{wpp}_{\Tilde{d}}(\mathbb{P}_a)=\{\rho_{\Tilde{d}}\in\mathbb{N}_{n_\text{max}}|(a|_{C_{\mathbb{P}_{\Tilde{d}}=\{\rho_{\Tilde{d}}\}}})\subseteq (a|_{C_{\mathbb{P}_a}})\}
\end{equation}
where $a|_{C_{\mathbb{P}_a}}$ represents the set of $a$ behaviors under configuration $C_{\mathbb{P}_a}$, and 
$a|_{C_{\mathbb{P}_{\Tilde{d}}=\{\rho_{\Tilde{d}}\}}}$ represents $a$ behaviors under configuration $C_{\mathbb{P}_{\Tilde{d}}}$ where
$\mathbb{P}_{\Tilde{d}}=\{\rho_{\Tilde{d}}\}$.

\begin{figure}[h]
    \centering
    \includegraphics[width=0.47\textwidth]{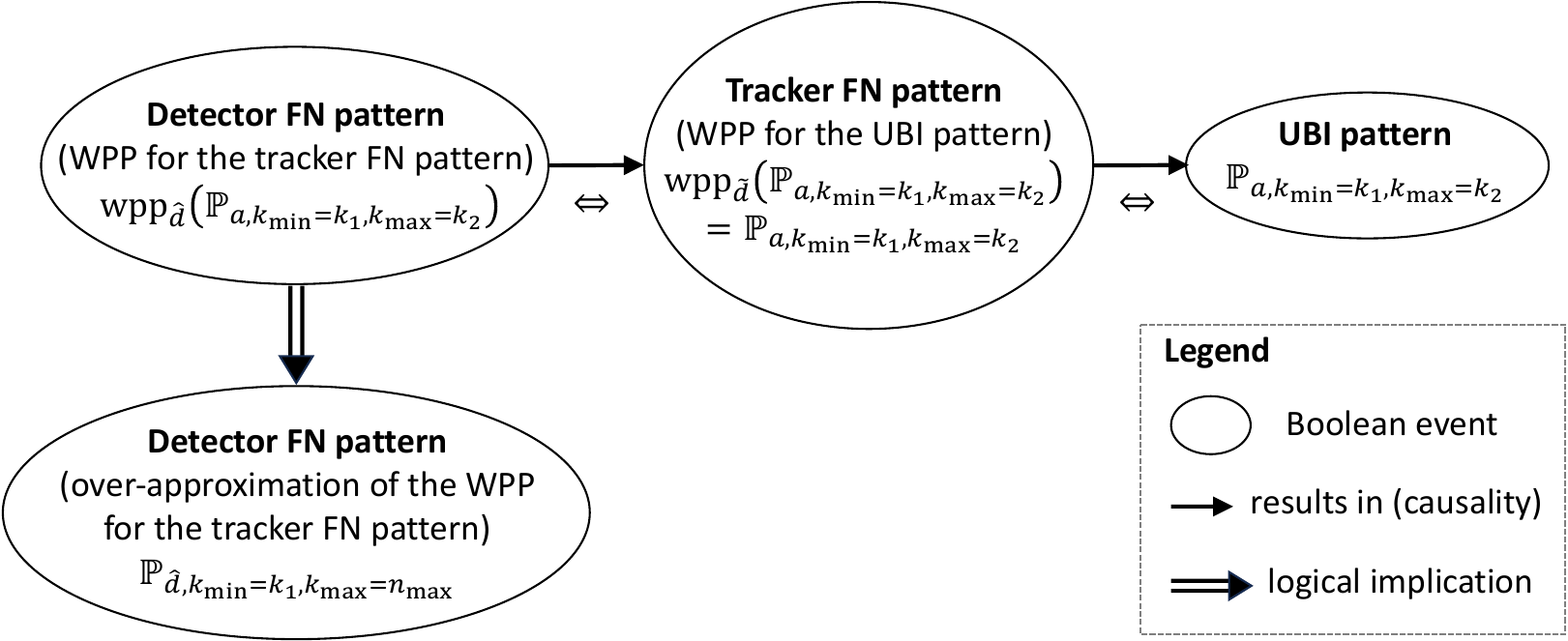}
    \caption{Causal model for \gls{UBI} pattern $\mathbb{P}_{a,k_\text{min}=k_1,k_\text{max}=k_2}$. Single-line arrows represent causality; double-line arrows represent logical implication (and no causality) and are used to model over-approximation. The bi-implication symbol $\Leftrightarrow$ annotating each causal arrow indicates, by a slight abuse of notation, an identity function associated with each of them, meaning that the pattern on the right of a causal arrow annotated with the bi-implication occurs if and only if the pattern on the left of the arrow occurs. Because of the direction of causality, the pattern on the right of a causal arrow will occur with some delay relative to the occurrence of the pattern on the left.}
    \label{fig:causal_model}
\end{figure}

The relationship between $\text{wpp}_{\Tilde{d}}(\mathbb{P}_a)$ and $\mathbb{P}_a$ can be seen as causal (see the right arrow in Fig.~\ref{fig:causal_model}): the occurrence of an \gls{FN} tracking error sequence from $\text{wpp}_{\Tilde{d}}(\mathbb{P}_a)$ causes the occurrence of a \gls{UBI} sequence from $\mathbb{P}_a$, and there are no other \gls{FN} sequences not captured in $\text{wpp}_{\Tilde{d}}(\mathbb{P}_a)$ that could cause a \gls{UBI} sequence in $\mathbb{P}_a$. Given a \gls{UBI} pattern $\mathbb{P}_{a,k_\text{min}=k_1,k_\text{max}=k_2}$, specified by bounding the number of time steps when \gls{UBI} occurs (eq.~\ref{eq:UBI-pattern}), its weakest precondition at $\Tilde{d}$ is the same pattern, i.e.,  $\mathbb{P}_{\Tilde{d},k_\text{min}=k_1,k_\text{max}=k_2}=\mathbb{P}_{a,k_\text{min}=k_1,k_\text{max}=k_2}=\text{wpp}_{\Tilde{d}}(\mathbb{P}_{a,k_\text{min}=k_1,k_\text{max}=k_2})$.
This is easy to see from the policy definition (eq.~\ref{eq:policy}), since injecting $r_\text{max}>s_\text{POV}$ at $\Tilde{d}$ (which is used as the first argument into $\pi(d,\dot{s})$) restricts the policy to its first two cases (i.e., $a=0$ or $a_\text{max}$, depending on speed; this is because $d>\Delta{s}_\text{stand}\land a_\text{b,req}<a_\text{b,min}$ is true under $d=r_\text{max}>s_\text{POV}$) and these are equivalent to injecting a \gls{UBI} at $a$. In other words, injecting an \gls{FN} error at $\Tilde{d}$ at step $k$ causes a \gls{UBI} at $a$ at that step. Note that, for simplicity, this analysis assumes perfect distance $\overline{d}$ and speed $\dot{s}$ estimation, as modeled in DSM' under $C_{\mathbb{P}_{\Tilde{d}}}$. The first one is based on the tracker taking $\overline{d}$ as input (see eq.~\ref{eq:tracker}), and the latter, which is the intended behavior of \textit{Odo}, is injected with $\gamma_{\dot{s}}=\mathbb{N}_{n_\text{max}}$. We also ignore any perception-reaction delays. We will discuss how to relax these assumptions in the next section.

The next step is to establish the cause of the \gls{FN} tracking error pattern $\mathbb{P}_{\Tilde{d},k_\text{min}=k_1,k_\text{max}=k_2}$, i.e., $\text{wpp}_{\hat{d}}(\mathbb{P}_{\Tilde{d},k_\text{min}=k_1,k_\text{max}=k_2})$, which is equivalent to $\text{wpp}_{\hat{d}}(\mathbb{P}_{a,k_\text{min}=k_1,k_\text{max}=k_2})$ (see the left causality arrow in Fig.~\ref{fig:causal_model}). Computing it requires analyzing the tracker logic (eq.~\ref{eq:tracker}), which compensates for $c$ consecutive \gls{FN} detection errors. Specifying $\text{wpp}_{\hat{d}}(\mathbb{P}_{a,k_\text{min}=k_1,k_\text{max}=k_2})$ is difficult, but we can easily provide a conservative over-approximation by observing that the tracker may reduce the number of \gls{FN} detection errors by its compensation logic but will never introduce additional ones. Also, if all \gls{FN} detection errors occur as a continuous sequence starting at $t=0$, there will be the same number of \gls{FN} tracking errors in the output. Thus, $\text{wpp}_{\hat{d}}(\mathbb{P}_{a,k_\text{min}=k_1,k_\text{max}=k_2})$ contains sequences each with at least $k_1$ \gls{FN} detection errors. Also, some sequences may have more than $k_2$ \gls{FN} detection errors, because of possible compensation by the tracker. Thus, we can over-approximate $\text{wpp}_{\hat{d}}(\mathbb{P}_{a,k_\text{min}=k_1,k_\text{max}=k_2})$ by $\mathbb{P}_{\hat{d},k_\text{min}=k_1,k_\text{max}=n_\text{max}}$, which is represented by the bottom node in Fig.~\ref{fig:causal_model}.

It is worth noting that the causal chain of the three events at the top of Fig.~\ref{fig:causal_model} can be understood as a \acrfull{SCM}~\cite{Pearl09}. An \gls{SCM} is a directed acyclic graph with nodes representing random variables, and arrows representing causal influence; the arrows incoming to a node are also associated with a function that maps the variables at the other end of the arrows to the node. In our case, the variables are Boolean and represent the occurrence of an error pattern, and the functions associated with the causal arrows are identity. The \gls{SCM} can also be understood as an \gls{FT}~\cite{Ruijters15}, which is a causal model with nodes as Boolean events; causal arrows associated with Boolean functions; the sink node (aka top event) representing a system-level failure; and its ancestors (in the causal direction) being errors or faults or both. In our case, the top event is an \gls{UBI} pattern and the other events represent error patterns. The fourth node at the bottom of the figure and the logical implication (which does not express causality) are not part of the \gls{SCM} concept, but allow us to express over-approximation, which is useful when the exact cause is difficult to specify. 

\begin{table}[h]
    \centering
    \begin{tabular}{|p{0.04\textwidth}|p{0.13\textwidth}|p{0.22\textwidth}|} \hline 
         \textbf{\gls{UBI} pattern}& \textbf{Tracker \gls{FN} pattern (at $\Tilde{{d}}$) being a \gls{WPP} of \gls{UBI}} & \textbf{Detector \gls{FN} pattern (at $\hat{{d}}$) being an over-approximation of \gls{WPP} of \gls{UBI}}\\ \hline 
         $\mathbb{P}_{a,\text{S0..3}}$& $\mathbb{P}_{\Tilde{{d}},\text{S0..3}}=\mathbb{P}_{a,\text{S0..3}}$ & ${\mathbb{P}_{\hat{{d}},\text{S0..3}}=} \mathbb{P}_{\hat{{d}},k_\text{min}=k_\text{contact},k_\text{max}=n_\text{max}}$\\ \hline 
         $\mathbb{P}_{a,\text{S1..3}}$& $\mathbb{P}_{\Tilde{{d}},\text{S1..3}}=\mathbb{P}_{a,\text{S1..3}}$ & ${\mathbb{P}_{\hat{{d}},\text{S1..3}}=} \mathbb{P}_{\hat{{d}},k_\text{min}=k_\text{S0},k_\text{max}=n_\text{max}}$\\ \hline 
         $\mathbb{P}_{a,\text{S2..3}}$& $\mathbb{P}_{\Tilde{{d}},\text{S2..3}}=\mathbb{P}_{a,\text{S2..3}}$ & ${\mathbb{P}_{\hat{{d}},\text{S2..3}}=} \mathbb{P}_{\hat{{d}},k_\text{min}=k_\text{S1},k_\text{max}=n_\text{max}}$\\ \hline
         $\mathbb{P}_{a,\text{S3}}$& $\mathbb{P}_{\Tilde{{d}},\text{S3}}=\mathbb{P}_{a,\text{S3}}$ & ${\mathbb{P}_{\hat{{d}},\text{S3}}=} \mathbb{P}_{\hat{{d}},k_\text{min}=k_\text{S2},k_\text{max}=n_\text{max}}$\\ \hline         
    \end{tabular}
    \caption{Tracker \gls{FN} patterns that cause and detector \gls{FN} patterns that may cause \gls{UBI} patterns of a given severity} 
    \label{tab:HEPs}
\end{table}

The causal model in Fig.~\ref{fig:causal_model} allows us to limit the occurrence probability of a \gls{UBI} of at least a given severity by limiting the occurrence of the \gls{FN} detection error pattern that represents the over-approximation of the cause of the \gls{UBI} pattern. This is because $P(\mathbb{P}_{\hat{d},k_\text{min}=k_1,k_\text{max}=n_\text{max}})\geq\mathbb{P}_{a,k_\text{min}=k_1,k_\text{max}=k_2}$ due to the over-approximation, and we can also set $k_2=n_\text{max}$ without affecting this inequality. Table~\ref{tab:HEPs} shows such \glspl{HEP} at $\Tilde{{d}}$ and $\hat{{d}}$ that cause or may cause, respectively, an \gls{UBI} pattern of the corresponding severity range.

In our sample scenario (Fig.~\ref{fig:nominal_braking}), injecting $\mathbb{P}_{\hat{{d}},\text{S0..3}}$ corresponds to the occurrence of $k_\text{min}$=19 or more \gls{FN} detection errors within 150 frames, which may cause 19 or more time steps with \gls{UBI} within the 150 time steps, and thus a potential collision. Conversely, no collision due to \gls{UBI} caused by \gls{FN} detection errors is possible in our scenario if fewer than 19 \glspl{FN} occur within 150 frames. Similarly, no collision of severity S3 due to \gls{UBI} caused by \gls{FN} detection errors is possible in our scenario if fewer than 35 \glspl{FN} occur within 150 frames.

Limiting the occurrence probability of the patterns in the third column in Tab.~\ref{tab:HEPs} allows us to limit the occurrence probability of the \gls{UBI} patterns with the corresponding severity in the first column, and thus limit the risk of harm due to \gls{UBI}. 

\subsubsection{\gls{IRC} Identification and \gls{HEP} Likelihood Estimation for \gls{AI}-based Components}
\label{subsubsec:IRCs}

The analysis in the previous subsection involved computing \glspl{WPP} for the conventional components in our example, i.e., the driving policy and the tracker, whose scenario-specific behavior can be modeled precisely and whose implementation in conventional software can be verified against the model by the methods prescribed in ISO 26262:2018~\cite{ISO26262}, including inspection. It is typically infeasible to produce such behavior specifications for \gls{AI}-based components that rely on deep neural networks, such as the object detector, because they often implement very complex functions over highly dimensional inputs, such as images. Additionally, the logic implemented by neural networks cannot be inspected the way conventional programs can (for a comprehensive discussion of these issues in the context of safety assurance, see \cite{salay2018using}). Therefore, \gls{AI}-based components require a different approach. 

Rather than attempting to determine \glspl{WPP} over inputs like images that would cause specific \glspl{HEP} on the output of an \gls{AI}-based component, the combination of sensor and detector is modeled probabilistically, that is, $p(\hat{{d}}|y')$ in our example, and the probability of \glspl{HEP} is estimated via testing. Testing requires test data that adequately covers the input conditions fed into an \gls{AI}-based component. For a perception component, these are the \glspl{IRC} reflecting $y'$ in $p(\hat{{d}}|y')$. While \glspl{IRC} like object and scenery appearance and weather conditions are multidimensional and complex, existing road ontologies (e.g., \cite{ISO34503,wisedrive}) can be used to express \glspl{IRC} and partition the range of \glspl{IRC} expected in operation. The test data, such as sensor recordings from drives in the target \gls{ODD}, potentially augmented with synthetic data, would then be used to estimate the \gls{HEP} probability in each partition. In our example, the \glspl{IRC} would include different types of vehicles as \gls{POV}, different poses, road configurations, and weather conditions. The test data would be used to estimate the probability of detector \glspl{HEP} (from Tab.~\ref{tab:HEPs}) under different \glspl{IRC}. In our example, the modeled scenario would represent one of such partitions, and we would estimate $P(\mathbb{P}_{a,\text{S1..3}}|y')$. The \gls{HEP} probabilities then need to be aggregated over the corresponding \glspl{IRC} and \glspl{HBSC} and their occurrence rates in operation to estimate the final safety risk due to \gls{UBI} in operation. In our example, the occurrence rate of $\mathbb{P}_{a,\text{S1..3}}$ due to the analyzed scenario conditions in operation would be $\lambda(\mathbb{P}_{a,\text{S1..3}},y')=P(\mathbb{P}_{a,\text{S1..3}}|y')\lambda(y')$, where $y'$ covers the combination of \glspl{IRC} and \glspl{HBSC}, and $\lambda(y')$ denotes their occurrence rate (such as per kilometer driven; see~\cite{deGelder19,J2980} for guidance on calculating exposure). The pattern occurrence rates would be then aggregated over the remaining scenario partitions (i.e., combinations of \glspl{IRC} and \glspl{HBSC}). This partitioning of \glspl{HBSC} and \glspl{IRC} and aggregation of probabilities and occurrence rates can be captured in a safety case, as described later.

In contrast to the approach outlined above, the \gls{SOTIF} standard asks for the identification of triggering conditions, which may neither be feasible nor necessary to assure safety.
The standard defines triggering conditions as those scenario conditions that cause \glspl{HB}; thus, they are the subset of \glspl{IRC} that cause \glspl{HEP}. Whereas knowing the causal link between \glspl{IRC} and \glspl{HEP} may benefit addressing the corresponding functional insufficiency and creating stronger safety cases, establishing this link may be difficult or even infeasible. For example, a misdetection of an object in an image may be triggered by the context of the object rather than the object itself; it may also be due to particular noise or general appearance of the scene that cannot be easily specified or described in words. Further, creating counterfactual images that change just the specific triggering condition while leaving all other conditions unchanged, which may be necessary to establish causality, may be infeasible. On the other hand, statistical causal influences may be more practical to establish, e.g., by injecting specific weather conditions into a data set and relating it to \gls{HEP} rate, but this approach may not be feasible for all types of \glspl{IRC}. It is worth noting that identifying triggering conditions for specific decision of an \gls{AI}-based component is the subject of Explainable \gls{AI}~\cite{XAIforAD}, but existing research still lacks explanation methods that would benefit safety assurance of \gls{DAS} in practice. Most importantly, establishing causal links between \glspl{IRC} and \glspl{HEP} does not seem required for assuring safety. Partitioning \glspl{IRC} and estimating HEP rates in each partitions can provide a statistical assurance~\cite{ISCaP} without the need for establishing causality. Lists of known common triggering conditions may still be used to inform the partitioning of \glspl{IRC} and test data selection. 

Guidance on assuring that \gls{AI}-based components meet their safety-related performance requirements, such as \gls{HEP} rates derived using \gls{MoSAFE}, is subject to ISO/PAS 8800~\cite{ISO8800}. 
The standard covers safety-related properties and risk factors impacting the insufficient performance and malfunctioning behavior of \gls{AI} systems and components. Most related to \gls{IRC} identification and \gls{HEP} rate estimation, it provides guidance on the derivation of suitable safety requirements as well as data quality and completeness requirements. It also provides guidance on measures for the control and mitigation of failures in \gls{AI} systems and components.

When the estimated \gls{HEP} occurrence rates for an \gls{AI}-based component exceed the upper limits imposed by the acceptance criteria, the \gls{DAS} needs to be modified to reduce the risk (Clause 8). The possible modifications include improving the performance of the \gls{AI}-based component to reduce the \gls{HEP} occurrence rates, robustifying the downstream components to be less sensitive to the error sequences in the HEPs, such as by increasing safety margins (e.g., see~
\cite{PURSS,Kobayashi21}), and restricting the \gls{ODD} to exclude the \glspl{IRC} that cause the high \gls{HEP} occurrence rates.

%% file: limitations.tex
\section{Additional Considerations and Future Work}
\label{sec:limitations}

The previous sections demonstrated \gls{MoSAFE} on a necessarily simplified example. This section discusses how to handle more complex scenarios and models, including multiple types of \glspl{HB} and multi-input errors, relaxing some of the assumptions made earlier. It also illustrates how to use \gls{MoSAFE} results as part of a safety case. Finally, it discusses challenges, limitations, and suggestions for future work.

\subsubsection{Designing Adequate Environment and System Models}
\gls{MoSAFE} relies on models of the \gls{DAS} and the road and vehicle environment to identify and evaluate \glspl{HBP} (Clause 6) and \glspl{HEP} (Clause 7), and the adequacy of these models determines the validity of the risk analysis results. The main choices relate to the level of abstraction, such as selection of the real-world phenomena and their approximations to be included. The \glspl{HLSM} represent the \glspl{HBSC} and driving policies thereunder and can be typically represented using simple kinematic models and rule-based policies, as in our example of braking for a stationary vehicle ahead. Existing safety frameworks provide examples of such kinematic and rule-based models for a wide range of scenarios~\cite{shalevshwartz2018formal,Hasuo23RSS}. The selection of \glspl{HBSC} can rely on existing road environment ontologies~\cite{ISO34503,wisedrive}. The \glspl{DSM} need to refine the input space into the \gls{DAS} by adding \glspl{IRC}. The refinement of this input space can again rely on the existing road environment ontologies, which provide standard classifications of road users and other objects, road structures, and weather and visibility conditions. The refined \gls{DAS} model includes the decision logic relevant to the scenario, which can be extracted from the detailed \gls{DAS} design. While our \gls{DAS} model ignored time delays, perception and reaction time delays should be included. In our example, this would be done by starting the scenario before the braking needs to occur, extending the \glspl{UBI} by reaction delay, and also including velocity and distance prediction logic in the model. The models should be verified against the \gls{DAS}, for example, using simulation testing.


\subsubsection{Misbehavior Injection}

The \gls{UBI} used in our example is applicable to a large range of scenarios that require braking as intended behavior.  To evaluate a wider applicability of our approach, we have modeled \glspl{UBI} occurring in braking for a stationary object ahead, braking for a slower vehicle merging in front, and braking for a hard-stopping front vehicle. We have also modeled \glsfirst{UA} when waiting behind a stopped vehicle and steady following of a front vehicle, and \glsfirst{FTY} at a stop-controlled intersection. We were able to model these scenarios and \glspl{HB} using the same approach as in the presented example. In these scenarios, in addition to \glspl{HBP} of the form $k_\text{min}..k_\text{max}$ \gls{HB} occurrences within $n_\text{max}$ time steps, we have also identified two more useful forms: $k_\text{min}..k_\text{max}$ consecutive \gls{HB} occurrences within $n_\text{max}$ time steps, and $k_\text{min}..k_\text{max}$ consecutive \gls{HB} occurrences at the beginning of a scenario. Future research should explore \glspl{HB} representing deviations from intended lateral behavior.

Another type of \gls{HB} applicable to our example is unintended insufficient braking (\gls{UIB}), which is braking with a lower level than intended (Tab.~\ref{tab:hazards}). In our example, we can define it as a reduction of the required braking $a_{b,\text{req}}$ by the factor $(1-\eta_{a_{b,\text{req}}})$, where is $\eta_{a_{b,\text{req}}}:[0..T_\text{max}]\rightarrow[0..1]$ is an error signal representing the relative reduction of $a_{b,\text{req}}$ over time. Note that an error signal is the continuous analogue of an error sequence. Similar as for \gls{UBI}, we can define a \gls{UIB} pattern by bounding the magnitude of the reduction by some maximum $0<\hat{\eta}\leq 1$:
\begin{equation}
    \mathbb{P}_{a_{b,\text{req}},\hat{\eta}}=\{\eta_{a_{b,\text{req}}}:[0..T_\text{max}]\rightarrow[0..\hat{\eta}]\}
    \label{eq:pred}
\end{equation}
Given a signal error $\eta_{a_{b,\text{req}}}\in\mathbb{P}_{a_{b,\text{req}},\hat{\eta}}$, it can be injected into \gls{HLSM} and \gls{DSM} by multiplying the the right-hand side of eq.~\ref{eq:areq} by $(1-\eta_{a_{b,\text{req}}})$.

We can analyze the worst-case effect of $\eta_{a_{b,\text{req}}}\in\mathbb{P}_{a_{b,\text{req}},\hat{\eta}}$ with the constant error signal with value $\hat{\eta}$. For example, Fig.~\ref{fig:reduced_braking} shows the speed and acceleration profile assuming a reduction  $\eta=0.14$ of the required braking $a_\text{b,req}$ throughout the scenario. For comparison, the dashed curves show the intended behavior of braking with $a_\text{b,req}$. As a result of the reduced braking, the \gls{SV} does not brake until $0.86a_\text{b,req}$ reaches $a_\text{b,min}$ at $s_\text{b,start}$; subsequently, the driving policy applying $0.86a_\text{b,req}$ compensates the initial lack of braking in order to stop for $s_\text{stop}$, requiring a high-level of deceleration before reaching it. For example, $0.86a_\text{b,req}$ reaches the maximum braking of 8\,$\text{m}/\text{s}^2$ about 2\,ms before stopping, but its speed is just 0.2\,$\text{m}/\text{s}$. As a result, the \gls{SV} will overshoot $s_\text{stop}$,  but only by less than 2.5\,cm, which is safe. Even a 50\,\% reduction of $a_\text{b,req}$ would lead to less than 2\,m overshoot, which would be safe assuming $\Delta s_\text{stand}=$5\,m.

The approach to represent and inject misbehaviors using binary or continuous patterns is quite general and flexible. Although our examples inject only constants and simple functions of the ground truth, more complex misbehaviors can also be injected.

\subsubsection{Fault Tree Derivation Under Multi-Input Failures}

The example in Fig.~\ref{fig:refined_model_inject} considered a single input into an element being erroneous at a time, such as a \gls{FN} track detection at $\Tilde{{d}}$ into the policy and the \gls{FN} object detection at $\hat{{d}}$ into the tracker. This resulted in the fault tree at the top of Fig.~\ref{fig:causal_model} being a sequence of three nodes. In practice, an element may receive more than one erroneous input at the same time. For example, the policy $\pi$ could receive both a pattern of \gls{FN} tracking errors at $\Tilde{{d}}$ and a pattern of erroneous speed estimates $\hat{\dot{s}}$ at the same time. One way to deal with this more general case is to partition the input error cases into \emph{single-input failure} and \emph{multi-input failure}, which will cause the fault tree to be a general tree rather than a sequence, and even more generally a directed acyclic graph if inputs are shared among multiple elements.

\begin{figure*}[h]
    \centering
    \includegraphics[width=0.8\textwidth]{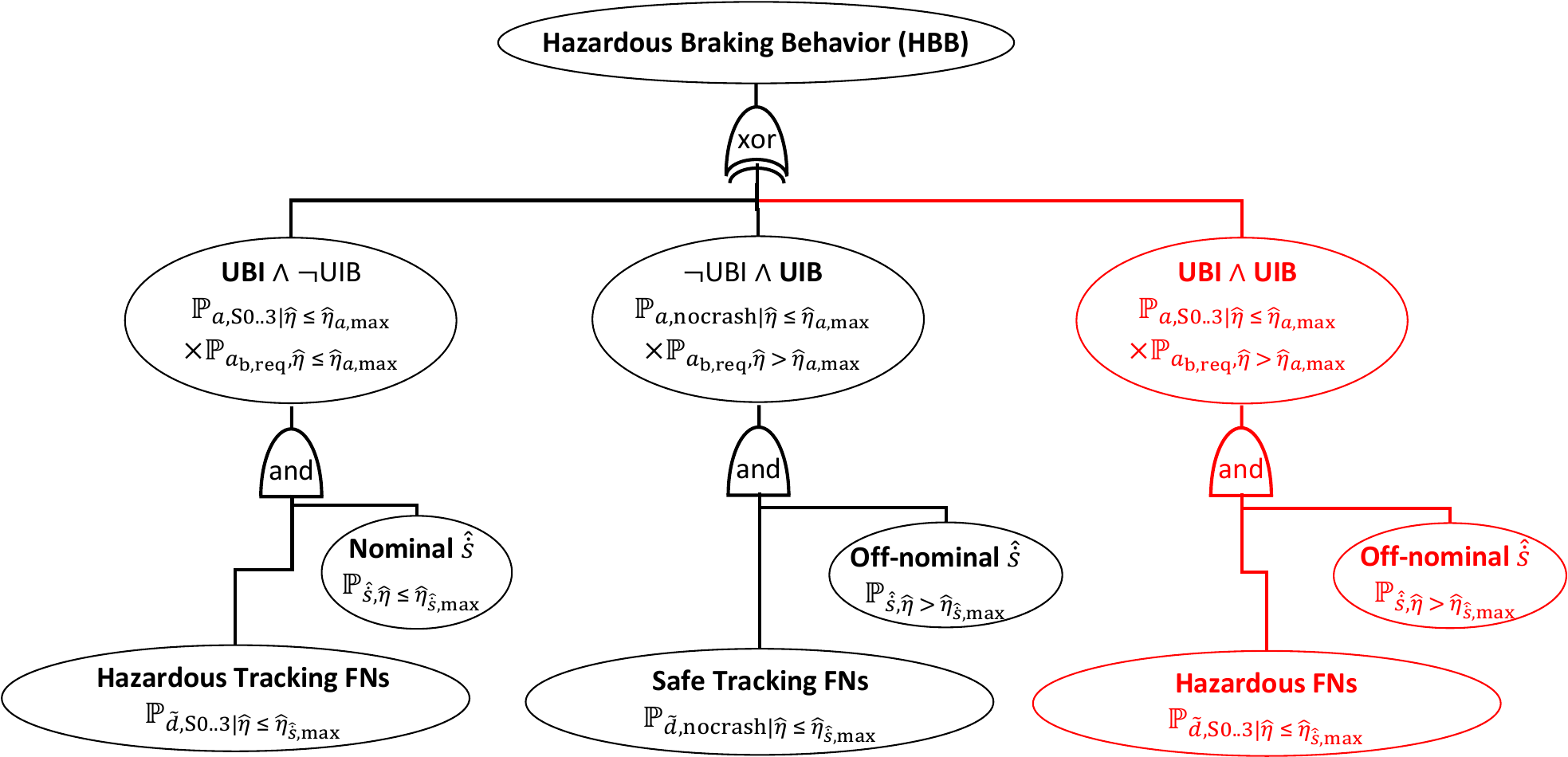}
    \caption{Sample fault tree considering multiple  inputs into an element being erroneous: hazardous braking behavior caused by \gls{FN} tracking errors or \gls{SV} speed underestimates or both}
    \label{fig:ft}
\end{figure*}

As an example, consider \gls{HBB} as a combination of hazardous \gls{UBI} and \gls{UIB}, ignoring \glsfirst{UHB} for simplicity. \gls{HBB} constitutes the top event of the fault tree in Fig.~\ref{fig:ft}. This event is partitioned into three cases:
\begin{enumerate}
    \item $\text{\gls{UBI}}\land\neg\text{\gls{UIB}}$: a hazardous \gls{UBI} with an otherwise nominal required braking level ($\neg\text{\gls{UIB}}$), i.e., one that is not reduced by more than some nominal maximum $\hat{\eta}_{a,\text{max}}$ when braking is not interrupted; this case is formalized as a combination of $\mathbb{P}_{a_{b,\text{req}},\hat{\eta}\leq\hat{\eta}_{a,\text{max}}}$ representing $\neg\text{\gls{UIB}}$ (eq.~\ref{eq:pred}) and $\mathbb{P}_{a,\text{S0..3}|\hat{\eta}\leq\hat{\eta}_{a,\text{max}}}$ representing \gls{UBI}, with the latter defined same as $\mathbb{P}_{a,\text{S0..S1}}$ in Tab.~\ref{tab:UBI}, but with $k_\text{contact}$ determined under  $a_{b,\text{req}}$ reduced according to $\mathbb{P}_{a_{b,\text{req}},\hat{\eta}\leq\hat{\eta}_{a,\text{max}}}$. Thus, the severity bounds for $\tau_\text{total}$ in Tab.~\ref{tab:severity} would need to be recomputed assuming the worst-case but still nominal reduction of $a_{b,\text{req}}$ by $\hat{\eta}_{a,\text{max}}$; this corresponds to \glspl{UBI} on a braking profile like in Fig.~\ref{fig:reduced_braking}. 
    
    \item $\neg\text{\gls{UBI}}\land\text{\gls{UIB}}$: a (potentially) hazardous \gls{UIB} that exceeds the nominal level $\hat{\eta}_{a,\text{max}}$, denoted by $\mathbb{P}_{a_{b,\text{req}},\hat{\eta}>\hat{\eta}_{a,\text{max}}}$, while \gls{UBI} is bounded to a level that would be safe under nominal braking reduction levels $\mathbb{P}_{a_{b,\text{req}},\hat{\eta}\leq\hat{\eta}_{a,\text{max}}}$, denoted by $\mathbb{P}_{a,\text{nocrash}|\hat{\eta}\leq\hat{\eta}_{a,\text{max}}}$;
    
    \item $\text{\gls{UBI}}\land\text{\gls{UIB}}$: a combination of hazardous \gls{UBI} and \gls{UIB}, i.e., $\mathbb{P}_{a,\text{S0..3}|\hat{\eta}\leq\hat{\eta}_{a,\text{max}}}\times\mathbb{P}_{a_{b,\text{req}},\hat{\eta}>\hat{\eta}_{a,\text{max}}}$. 
\end{enumerate}
Formally, each of the three composite patterns is a cross-product. 

Looking at eq.~\ref{eq:areq}, a \gls{UIB} can be caused by an underestimate of $\dot{s}$ or an overestimate of $d$ or both. For simplicity, we only consider underestimates of $\dot{s}$. Based on eq.~\ref{eq:areq}, injecting $\mathbb{P}_{\hat{\dot{s}},\hat{\eta}\leq\hat{\eta}_{\hat{\dot{s}},\text{max}}}$ with $\hat{\eta}_{\hat{\dot{s}},\text{max}}=1-\sqrt{1-\hat{\eta}_{a,\text{max}}}$ is equivalent to injecting $\mathbb{P}_{a_{b,\text{req}},\hat{\eta}\leq\hat{\eta}_{a,\text{max}}}$; similarly, injecting $\mathbb{P}_{\hat{\dot{s}},\hat{\eta}>\hat{\eta}_{\hat{\dot{s}},\text{max}}}$ is equivalent to injecting $\mathbb{P}_{a_{b,\text{req}},\hat{\eta}>\hat{\eta}_{a,\text{max}}}$. The patterns for $\hat{\dot{s}}$ are defined similarly to eq.~\ref{eq:pred} and injected by multiplying $\dot{s}$ in eq.~\ref{eq:areq} by $(1-\eta_{\hat{\dot{s}}})$.  

Each of the three cases under the top event are then decomposed using \textit{and}-gates (Fig.~\ref{fig:ft}). The patterns on $\Tilde{d}$ are defined as in the second column of Tab.~\ref{tab:UBI}, but with severity bounds determined by assuming nominal speed underestimates $\mathbb{P}_{\hat{\dot{s}},\hat{\eta}\leq\hat{\eta}_{\hat{\dot{s}},\text{max}}}$. The first two cases are examples of single-input failures, where one of the two inputs $\Tilde{d}$ and $\hat{\dot{s}}$ exceeds its threshold while the other does not, and the third case is a multi-input failure where both inputs exceed their thresholds.

The fault tree (Fig.~\ref{fig:ft}) allows computing the probability of the top event given the probability of the leaves (we use the descriptive names of the nodes rather then the patterns for readability):
\begin{equation}
    \begin{aligned}
        P(\text{\gls{HBB}})=\\
        P(\text{Hazardous-tracking-FNs}\land\text{Nominal-speed-estimate})+\\
        P(\text{Safe-tracking-FNs}\land\text{Off-nominal-speed-estimate})+\\
        P(\text{Hazardous-tracking-FNs}\land\text{Off-nominal-speed-estimate}) 
    \end{aligned}
\end{equation}
Since \glspl{FN} and \gls{SV} speed estimation errors are independent in our system, we have
\begin{equation}
    \begin{aligned}
        P(\text{\gls{HBB}})=\\
        P(\text{Hazardous-tracking-FNs})P(\text{Nominal-speed-estimate})+\\
        P(\text{Safe-tracking-FNs})P(\text{Off-nominal-speed-estimate})+\\
        P(\text{Hazardous-tracking-FNs})P(\text{Off-nominal-speed-estimate}) 
    \end{aligned}
\end{equation}
Finally, since both P(\text{Hazardous-tracking-FNs}) and P(\text{Off-nominal-speed-estimate}) are small, and thus the probabilities of their complements, P(\text{Safe-tracking-FNs}) and P(\text{Nominal-speed-estimate}), respectively, are close to one, we have
\begin{equation}
    \begin{aligned}
        P(\text{\gls{HBB}})\approx\\
        P(\text{Hazardous-tracking-FNs})+\\
        P(\text{Off-nominal-speed-estimate}) 
    \end{aligned}
\end{equation}
This modeling and probability-calculation approach, allowing to eliminate the multiple-failure branch marked in red in Fig.~\ref{fig:ft}, is standard in \acrfull{FTA}~\cite{Ruijters15,FTA}. Further, the approach of combining error patterns on multiple inputs can be easily extended to more than two erroneous inputs, which would allow us to also include a hazardous over-estimate of $\Tilde{d}$, $\mathbb{P}_{\Tilde{d},\hat{\eta}>\hat{\eta}_{\Tilde{d},\text{max}}}$, as another cause of \gls{UIB}, and a ``nominal'' over-estimate of $\Tilde{d}$, $\mathbb{P}_{\Tilde{d},\hat{\eta}\leq\hat{\eta}_{\Tilde{d},\text{max}}}$, for $\neg\text{\gls{UIB}}$.

In some cases, such as when modeling fault-tolerance of redundant elements, we want to define an \gls{HEP} over two or more inputs rather than freely composing \glspl{HEP} over individual inputs. Such \glspl{HEP} over multiple inputs would be useful to model simultaneous \gls{FN} detection errors in multiple sensor modalities, e.g., camera and lidar.

\subsubsection{\gls{WPP} Derivation}

The computation of \glspl{WPP} with \gls{FN} tracking error sequences causing hazardous \gls{UBI} patterns in our example was particularly simple due to the simple driving policy logic, but even the relatively simple tracker logic required us to resort to an over-approximation to express \glspl{WPP} with \gls{FN} detection error sequences that cause hazardous \gls{UBI} patterns. The scenario-specific decision logic of an element captured by the element model, although simplified compared to the full decision logic of the element, may still cause the \gls{WPP} computation to be challenging. The computation should be supported by reasoning tools, which should be explored in future work. In particular, the tooling should allow engineers to strike a balance between the complexity of a \gls{WPP} and its precision.

A causal model may contain multiple over-approximation nodes and implication arrows. \gls{WPP} computation may be applied to an over-approximation, and the \gls{WPP} itself might need to be over-approximated again. Thus, the ability to mix causal arrows and implications in causal models is important.


\subsubsection{Model Validation Using Simulation Testing}
\label{sec:sim}
Testing is also necessary to validate the \gls{DAS} and \gls{SV} models used in \gls{MoSAFE}, that is, assuring that they reflect the actual behavior of the \gls{DAS} and the \gls{SV} with sufficient accuracy. This approach of targeted testing to validate assumptions follows the layered residual-risk framework to \gls{ADS} testing~\cite{koopmantoward}. One of the basic ideas of this framework is to ``use higher-fidelity simulation and tests to reduce residual risks due to assumptions and gaps in lower fidelity simulations and tests.''~\cite{koopmantoward} Similarly, closed-course tests can be used to reduce residual risks due to assumptions and gaps in simulation testing. Finally, field testing is used primarily to discover new scenarios and requirements, including gaps in scenario catalogs, and to collect field data for testing \gls{AI}-based components and for estimating exposure, i.e., occurrence rates of scenario conditions. 

For our sample braking scenario, we have compared the analytical results of \gls{MoSAFE} from the \gls{HLSM} and \gls{DSM} with the test results of injecting \glspl{HEP} into WISE-ADS, a real \gls{ADS} software stack running in a high-fidelity simulator~\cite{Antkiewicz20}. The \gls{ADS} uses a lidar-based object detector using a deep neural network, a multi-object tracker based on a Kalman filter, a rule-based behavior planner, an optimization-based trajectory planner, a PID speed controller and a geometric steering controller, implemented on top of the Robot Operating System. In particular, the testing involved injecting a single sequence of consecutive \gls{FN} detection errors of various lengths and at different points in time during the braking scenario by intercepting the communication between the object detector and the tracker. Based on the dynamics represented by the \gls{HLSM}, we know that only consecutive \glspl{FN} provide the largest risk effect, i.e., the smallest slowdown, for a given number of \glspl{FN}, and thus patterns of intermittent \glspl{FN} do not need to be simulated for this scenario. The simulation environment includes a high-fidelity 14-\glsdisp{DOF}{DOF} vehicle dynamics model of the physical \gls{SV}~\cite{VanGennip18,Hosking18}. The braking behavior of the \gls{ADS} and the resulting velocity profiles in simulation were experimentally confirmed to closely resemble the behavior of the \gls{SV}, 2015 Lincoln MKZ, controlled by the ADS on a test track~\cite{Antkiewicz20}.

The \gls{DAS} model in Fig.~\ref{fig:refined_model} reflects the main components of the \gls{ADS} that are involved in the braking scenario. The model was updated to account for perception-reaction time, including the transition from maximum acceleration the maximum braking (such as at the end of the second \gls{UBI} interval in Fig.~\ref{fig:multi_interrupted_braking}), effectively extending the duration of the \gls{UBI} with $a_\text{max}$ before $a_\text{b,max}$ is applied. The \gls{DSM} parameters (i.e., $a_\text{b,min}$, $a_\text{b,max}$, $a_\text{max}$, $v_\text{max}$, $\Delta s_\text{stand}$, $\Delta t$, and $c$ in Fig.~\ref{fig:refined_model} and the perception-reaction delays) were calibrated with parameter values matching those of the \gls{ADS} and \gls{SV} combination, based on simulation experiments. Given these parameter values, \gls{WPP} analysis as presented earlier predicts that injecting 7 or fewer frames with an \gls{FN} detection error for the stationary \gls{POV} during the scenario should be safe (this injection corresponds to $C_{\mathbb{P}_{\hat{d}}}$ in the last row in Tab.~\ref{tab:configs}). If the \gls{POV} has been detected at the beginning of the scenario, then this number increases by $c=9$ frames, which are compensated by the tracker. Since our \gls{DSM} predicts that injecting 7 or fewer frames with an \gls{FN} detection error for the  \gls{POV} at the beginning of the scenario is safe, we determine that injecting $k$=7+9=16 or fewer frames with an \gls{FN} detection error for the \gls{POV} is safe.

To assess the result from \gls{MoSAFE}, we run a simulation experiment of injecting $k$=23, 24, 25, 26, 27 consecutive frames with \gls{FN} detection error for the \gls{POV} into the tracker in the actual ADS running in the high-fidelity simulator, with the injection occurring at the optimal point during the scenario as determined by the calibrated \gls{DSM}, and repeating the simulation 100 times for each $k$. We observed that $k$=23 or less was safe: $k$=23 resulted in 0 collisions, whereas $k$=24 resulted in 17\,\% of the runs, and $k$=27 resulted in 100\,\% of the runs experiencing a collision. The estimate from the \gls{DSM} of $k$=16 or less being safe is lower than $k$=23 from simulation. This is mainly because the \gls{DSM} makes the conservative assumption that the \gls{SV} accelerates with $a_\text{max}$ during the \gls{UBI}, but the \gls{SV} accelerated at lower rates than the maximum in the actual simulation runs. Given that the \gls{SV} with no \gls{UBI} stops within $n_\text{max}$=56 frames, the crash risk from \gls{UBI} in this scenario can be limited by limiting the probability of $k$=24 or more frames with \gls{FN} detection error for the \gls{POV} occurring anywhere (including intermittently) within  $n_\text{max}$=56 consecutive frames.

Whereas simulation testing is still required to validate the \gls{DSM}, the validation is much less costly than exploring the effect of injecting error sequences without a model. The analytical model states that multi-interval \gls{UBI} can always be replaced by a single \gls{UBI} interval with the same duration and a more severe effect, and thus the main uncertainty to be addressed by targeted simulation is about the system transition dynamics, the actual acceleration and deceleration profiles, and the \gls{SV}-speed and distance-to-POV estimation errors. Running the 500 simulations in our validation experiment (i.e., 100 runs for each of the five values of $k$) took 7.5\,h on a single modern desktop computer. Exploration of the effects of injecting \gls{FN} detection error sequences would have likely required orders of magnitude more simulation runs with random error sequences injected to arrive at similar results, which would have been prohibitively expensive.

\subsubsection{Trade-offs Between Model-Based Verification and Simulation Testing}

\gls{MoSAFE} can be viewed as model-based, modular verification, and an alternative to this approach is to perform \emph{automated black-box testing} in simulation~\cite{Corso22}. Such testing samples inputs automatically and executes the system under test as a black-box in a simulation environment. To be effective, it needs to use some form of optimization during input sampling in order to focus on finding and exploring hazardous inputs efficiently. These automated approaches can be divided into three categories: (i) falsification, which finds inputs that violate a safety property, i.e., cause an \gls{HB}; (ii) most-likely failure analysis, which tries to find maximum-likelihood failures, i.e., most-likely \gls{HB} occurrences; and (iii) failure probability estimation, which estimates the probability of \gls{HB} occurrence~\cite{Corso22}. Approaches in the third category are most relevant to risk estimation.

Black-box testing can be applied to a whole \gls{DAS} or parts of it. For example, it could be used to test and evaluate risk by generating perception inputs into a \gls{DAS}, either by generating them using computer graphics or synthesizing them using neural rendering (e.g., neural radial fields~\cite{NSG21}), or synthetically perturbing real or synthetic inputs, e.g., by adding weather effects. Alternatively, it can be applied to the prediction and planning portion by using simulated perception results as input. Black-box testing can also be applied in a gray-box setting by injecting error sequences into internal interfaces of the \gls{DAS}, for example, to understand how prediction errors might propagate through the system (e.g., \cite{Jha19}), and to leverage the test optimization capability to find component inputs that cause \glspl{HB}.

There are important trade-offs between model-based verification and simulation-based black-box testing.
The key advantage of black-box testing methods is that it can be performed on the actual system software implementation and does not require a model of the system. On the other hand, while model-based verification requires the additional effort to create models, it can provide stronger guarantees than black-box testing, because it performs an exhaustive analysis of the models. In particular, an \gls{HLSM} represents an exponentially large class of concrete scenarios, because it is parameterized, and MoSAFE analyzes it exhaustively. Further, the assume-guarantee reasoning over a \gls{DSM}, as in the \gls{WPP} derivation, establishes the full causal links between errors and system failures, which can be leveraged in a safety case. Black-box testing may require large number of samples to generate insights comparable to verification, and it does not provide guarantees. Also, having these \glspl{WPP} as part of interface contracts allows modular verification, including modular testing. For example, \gls{AI} components can be subjected to unit testing against an interface specifying \gls{HEP} occurrence rate limits, which enables independent development, as in the case of automotive supply chains, and it also can improve test depth given the same amount of test budget~\cite{shalev2016sample}. 

Ultimately, it may be best to combine both approaches in practice. At the vehicle level, kinematic models of the road-and-vehicle environment and simple rule-based driving policies often allow for closed-form specification of \gls{HB} patterns, as in our example. More complex scenarios and policies may require black-box test simulation, with a potentially significant computational cost and loss of guarantees. Similarly, element-level models of \gls{DAS} may include complex logic necessitating black-box testing, possibly by injecting error sequences at element level. One may also combine black-box testing with specification mining to approximate \glspl{WPP}~\cite{Bartocci22}. Finally, simulation testing is required to validate \gls{DAS} and \gls{RVE} models, as already discussed.

\subsubsection{Using \gls{MoSAFE} Results in a Safety Case}

\gls{MoSAFE} results can be used as evidence in a \gls{DAS} safety case. A \gls{DAS} safety case is a claim supported by a structured argument founded on evidence that justifies the \gls{DAS} is considered to be safe in its operation~\cite{ISO5083}. We illustrate the use of \gls{MoSAFE} results in a safety case template for an \gls{ADS}. A safety case template provides a reusable argument structure, and our example is based on \gls{ISCaP}~\cite{ISCaP}. Whereas \gls{ISCaP} focuses on assuring perception components, the template in this paper generalizes to any type of components and it also includes vehicle and system-level arguments. 

\begin{figure*}[h]
    \centering
    \includegraphics[width=0.8\textwidth]{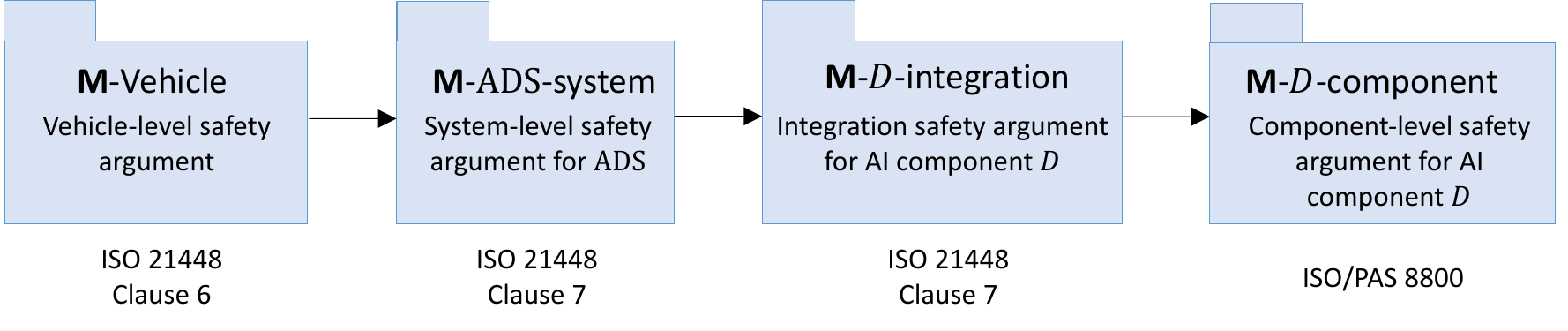}
    \caption{Safety case modules structured by the system decomposition. \gls{MoSAFE} addresses vehicle-level, system-level, and integration safety arguments.}
    \label{fig:gsn-modules}
\end{figure*}

The safety case template is organized into modules, reflecting the different levels of system decomposition (Fig.~\ref{fig:gsn-modules}). The top module, \textbf{M}-Vehicle, represents the vehicle level. The subsequent level focuses on the vehicle systems; although Fig.~\ref{fig:gsn-modules} only displays \textbf{M}-ADS-System for the \gls{ADS}, other modules at this level would encompass the remaining vehicle systems, including power train, braking, and electrical-power supply. The following level addresses the integration of components into their respective systems, exemplified by \textbf{M}-$D$-integration, which pertains to the integration of the object detector $D$ into the \gls{ADS}. Lastly, the component-level deals with the safety requirements at the component level, such as \textbf{M}-$D$-component for the object detector. The vehicle level is governed by the \gls{SOTIF} Clause 6 and the \gls{HLSM} analysis in \gls{MoSAFE}. The system and integration levels are covered by the \gls{SOTIF} Clause 7 and the \gls{DSM} analysis in MoSAFE. Finally, the component-level for \gls{AI} components is addressed by ISO/PAS 8800~\cite{ISO8800}.

\begin{figure*}[h]
    \centering
    \includegraphics[width=0.8\textwidth]{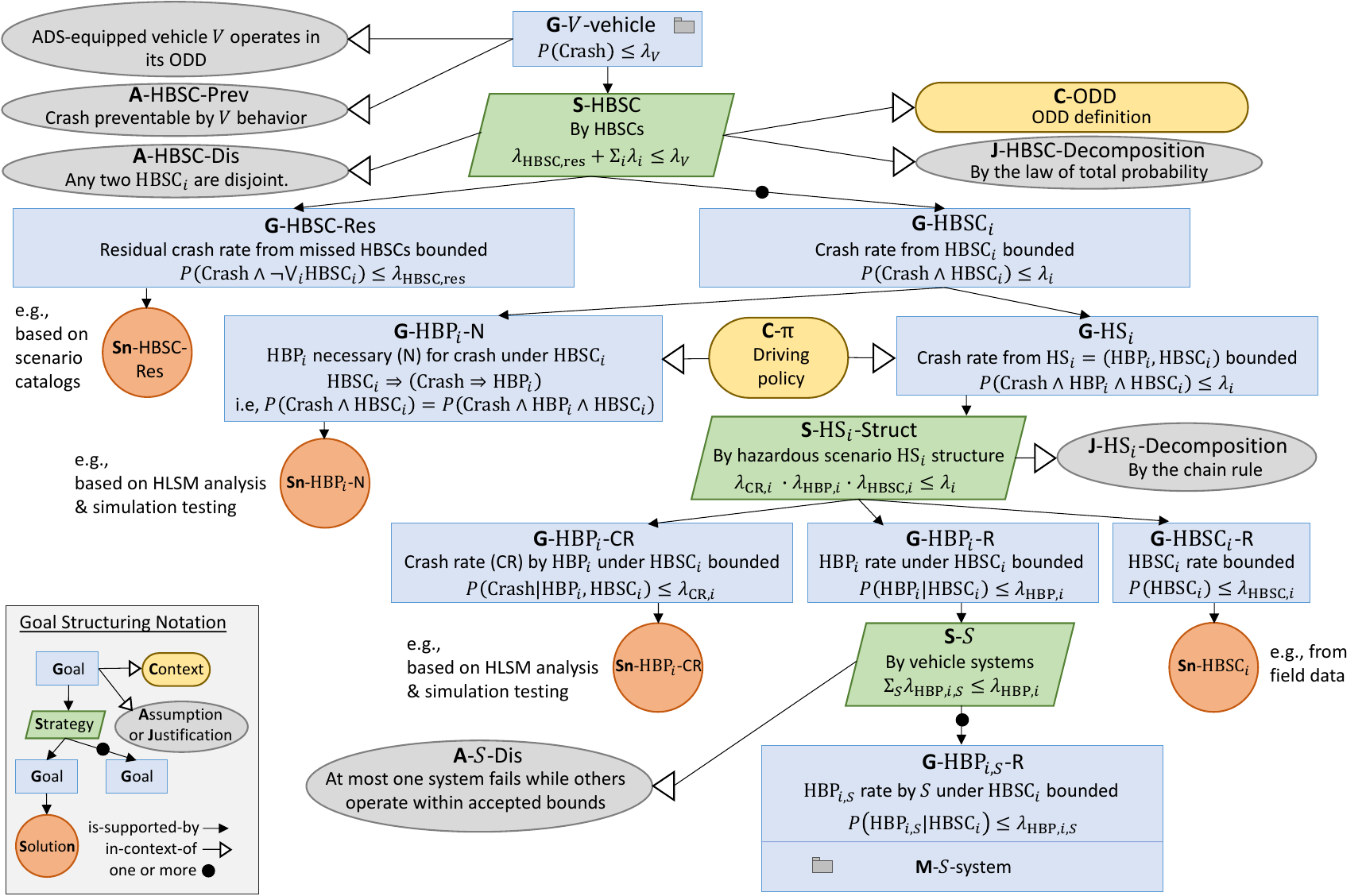}
    \caption{Vehicle-level safety argument template (i.e., content of the module \textbf{M}-Vehicle from Figure~\ref{fig:gsn-modules}) expressed in \gls{GSN}}
    \label{fig:vehicle-level-argument}
\end{figure*}

\begin{figure}[h]
    \centering
    \includegraphics[width=0.4\textwidth]{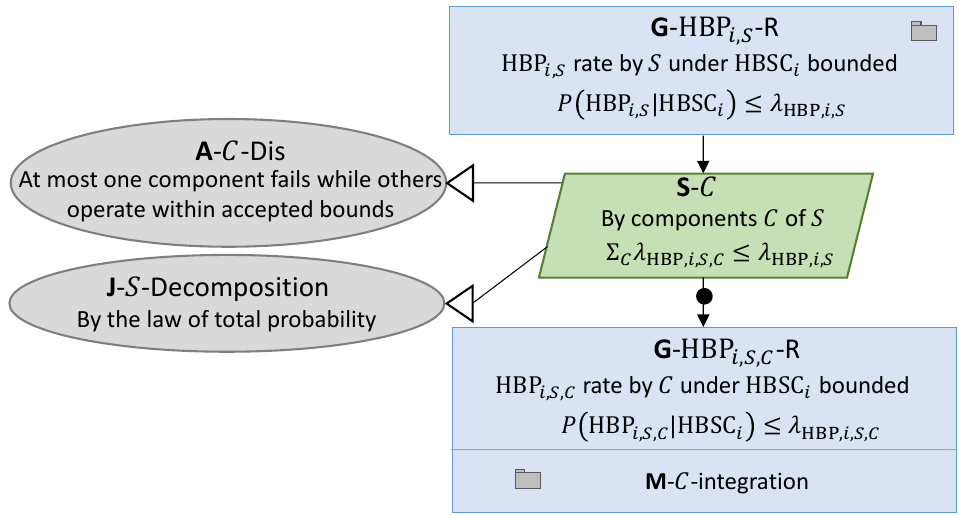}
    \caption{System-level safety argument template for system $S$ (i.e., content of the module \textbf{M}-ADS-system from Figure~\ref{fig:gsn-modules}, where $S$=ADS) expressed in \gls{GSN}}
    \label{fig:system-level-argument}
\end{figure}

\begin{figure*}[h]
    \centering
    \includegraphics[width=0.8\textwidth]{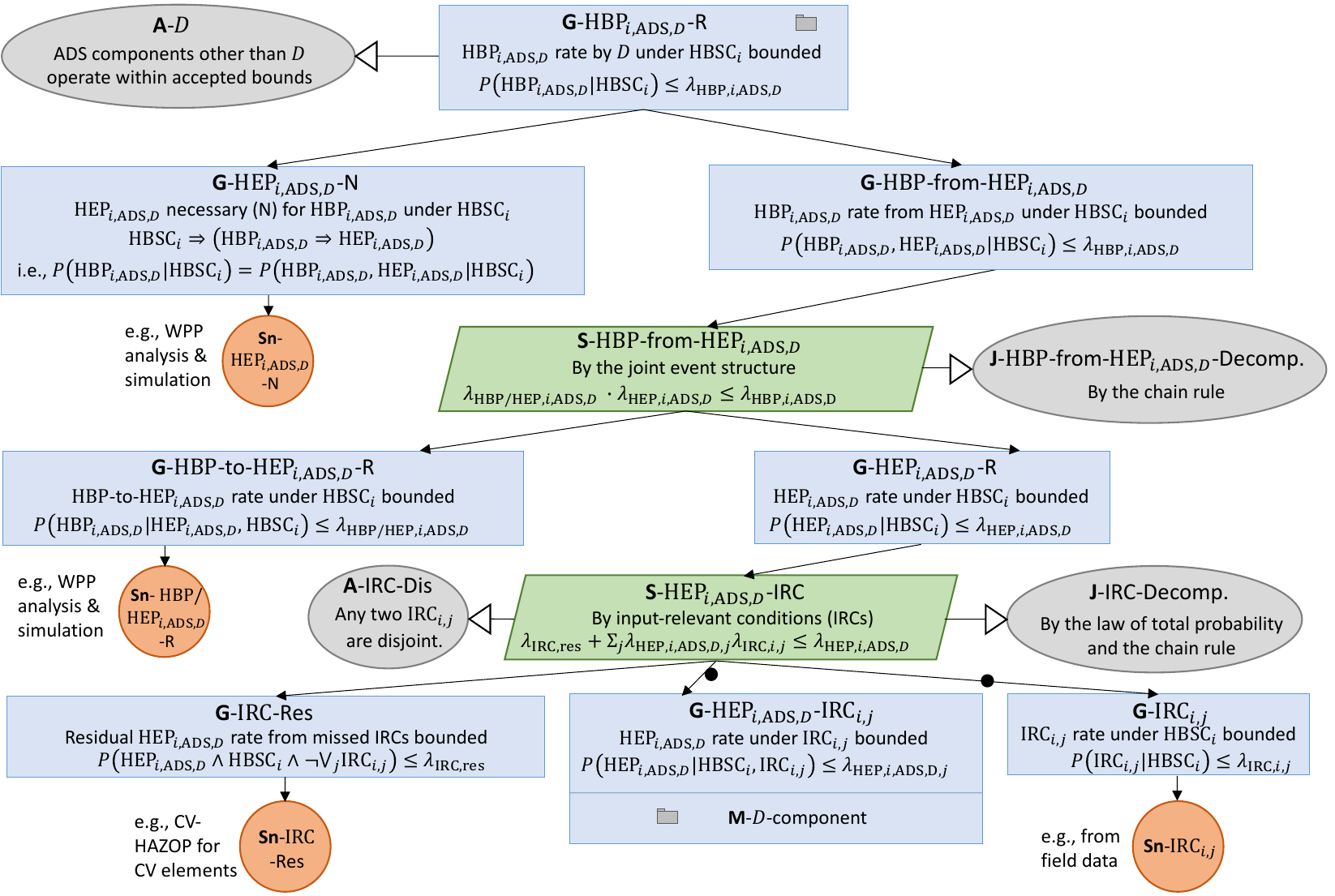}
    \caption{Integration safety argument template for the object detector $D$ (i.e., content of the module \textbf{M}-$D$-integration from Fig.~\ref{fig:gsn-modules}) expressed in \gls{GSN}}
    \label{fig:integration-argument}
\end{figure*}

The contents of the modules in the scope of \gls{MoSAFE} are detailed in Figs.~\ref{fig:vehicle-level-argument}-\ref{fig:integration-argument}. This template adopts a deductive reasoning approach for deconstructing the top-level claim, while confining any inductive reasoning processes, such as extrapolating from test results, to terminal claims that are substantiated directly by evidence~\cite{rushby2015interpretation}. The structure of the argument is represented using \acrfull{GSN}~\cite{GSNstandard} (see the legend in Fig.~\ref{fig:vehicle-level-argument} for a summary). In this notation, each goal node symbolizes a claim, and the strategy nodes elucidate how a higher-level goal is derived from its underlying supportive goals. Assumptions and justifications are articulated through oval-shaped nodes. Context nodes supply relevant contextual information, which includes additional documents and specifications necessary for interpreting the goals or strategies. Lastly, solution nodes cite references to the supporting evidence.

The content of the vehicle-level module, as illustrated in Fig.~\ref{fig:vehicle-level-argument}, centers on the top-level goal (\textbf{G}-V-vehicle), which claims that the overall crash rate attributable to the \gls{SV} behavior within the \gls{ODD} adheres to the upper limit $\lambda_v$ as a \gls{SOTIF} acceptance criterion. This criterion, $\lambda_v$, is established using principles such as \gls{GAMAB}, \gls{ALARP}, and \gls{MEM} as per \cite{ISO21448} and \cite{EN50126}. It is then distributed across the individual \glspl{HBSC} constituting the \gls{ODD}: $\lambda_{\text{HBSC,res}} + \sum_i\lambda_i\leq\lambda_v$. The black dot on the `is-supported-by' arrow denotes a multiplicity of $\text{HBCS}_i$. Assuming these $\text{HBCS}_i$ are mutually exclusive, this strategy can be understood mathematically as an application of the law of total probability: $P(\text{Crash})=P(\text{Crash}\wedge\neg \bigvee_i\text{HBCS}_i)+\sum_i P(\text{Crash}\wedge\text{HBCS}_i)$. The residual risk, $\lambda_{\text{HBSC,res}}$, accounts for any overlooked crash-relevant \glspl{HBSC}, and established scenario catalogs (e.g., ~\cite{najm2007pre,safetypool}) can be used to estimate this residual risk, as indicated by the solution node \textbf{Sn}-HBSC-Res. The crash rate for a specific $\text{HBSC}_i$, i.e., $P(\text{Crash}\wedge\text{HBCS}_i)$, is then estimated via its corresponding $\text{HBP}_i$, which fully encapsulates the hazardous behavior of the \gls{SV} under $\text{HBSC}_i$. Thus, we require that, under $\text{HBSC}_i$, $\text{HBP}_i$ is necessary for a crash: $\text{HBSC}_i\Rightarrow(\text{Crash}\Rightarrow \text{HBP}_i)$. This necessity claim, \textbf{G}-$\text{HBP}_i$-N, can be established using the \gls{HLSM} analysis in \gls{MoSAFE}, indicated by \textbf{Sn}-$\text{HBP}_i$-N. In our running example, the \gls{RVE} in Fig.~\ref{fig:high_level_model} and $\mathbb{P}_{a,\text{S0..3}}$ in Table~\ref{tab:UBI} represent sample $\text{HBSC}_i$ and $\text{HBP}_i$, respectively, assuming \gls{UBI} as the only \gls{HB} for this scenario. This necessity claim can be equivalently stated as $P(\text{Crash}\wedge\text{HBSC}_i)= P(\text{Crash}\wedge\text{HBP}_i \wedge\text{HBSC}_i)$. Consequently, the goal \textbf{G}-$\text{HBSC}_i$ can be restated as \textbf{G}-$\text{HS}_i$ (see Fig.~\ref{fig:vehicle-level-argument}), where $\text{HS}_i$ refers to the hazardous scenario combining $\text{HBP}_i$ and $\text{HBSC}_i$: $ P(\text{Crash}\wedge\text{HBP}_i \wedge\text{HBSC}_i)\leq\lambda_i$. This joint probability is then decomposed using the chain rule: $P(\text{Crash}\wedge\text{HBP}_i \wedge\text{HBSC}_i)=P(\text{Crash}|\text{HBP}_i ,\text{HBSC}_i)P(\text{HBP}_i |\text{HBSC}_i)P(\text{HBSC}_i)$. Establishing a tight bound on $P(\text{Crash}|\text{HBP}_i ,\text{HBSC}_i)$ is difficult, but it can be conservatively bounded by $\lambda_{{CR},i}=1$. The occurrence rate of $\text{HBSC}_i$, i.e., $P(\text{HBSC}_i)$, can be estimated from field data. Lastly, the $\text{HBP}_i$ occurrence rate under $\text{HBSC}_i$ is distributed over the vehicle systems, resulting in a set of $\lambda_{\text{HBP},i,S}$ in \textbf{G}-$\text{HBP}_{i,S}$-R, with each $\lambda_{\text{HBP},i,S}$ defining an $\text{HS}_i$-specific acceptance criterion for a given system $S$. The latter goal is then further decomposed in the module \textbf{M}-$S$-system over the components $C$ of $S$, as illustrated in Fig.~\ref{fig:system-level-argument}.

Figure~\ref{fig:integration-argument} shows the content of the integration module for the object detector $D$ as a component of the \gls{ADS}. As a first step, $P(\text{HBP}_{i,\text{ADS},D}|\text{HBSC}_i)$, which is the $\text{HBP}_i$ rate due to $D$ as part of the \gls{ADS} under $\text{HBSC}_i$, is over-approximated by $P(\text{HEP}_{i,\text{ADS},D}|\text{HBSC}_i)$, i.e., the rate of $\text{HEP}_{i,\text{ADS},D}$ under $\text{HBSC}_i$. This is done by showing $\text{HEP}_{i,\text{ADS},D}$ to be necessary for the $\text{HBP}_{i,\text{ADS},D}$ under $\text{HBSC}_i$ using the \gls{WPP} analysis of the \gls{DSM} for the $\text{HS}_i$ and assuming $\lambda_{\text{HBP/HEP},i,\text{ADS},D}=1$. In our running example, the $\text{HEP}_{i,\text{ADS},D}$ for causing $\mathbb{P}_{a,\text{S0..3}}$ as $\text{HBP}_{i,\text{ADS},D}$ under our sample scenario ``approach to a stationary \gls{POV}''  is $\mathbb{P}_{\hat{{d}},\text{S0..3}}$ from Tab.~\ref{tab:HEPs}. Thus, the goal of limiting $P(\text{HBP}_{i,\text{ADS},D}|\text{HBSC}_i)$ is realized by limiting $P(\text{HEP}_{i,\text{ADS},D}|\text{HBSC}_i)$. This is supported by the \gls{WPP} analysis of the \gls{DSM} in Figs.~\ref{fig:refined_model} and \ref{fig:causal_model} and the validation test results in simulation, as described earlier, and indicated by \textbf{Sn}-$\text{HEP}_{i,\text{ADS},D}$-N. For our sample approach scenario and the experimental \gls{SV} equipped with WISE-ADS, the analysis and simulations discussed earlier show that bounding the probability of $k_\text{contact}=24$ or more \glspl{FN} occurring within $n_\text{max}=56$ consecutive frames will bound the probability of a collisions due to \gls{UBI}. This probability $P(\text{HEP}_{i,\text{ADS},D}|\text{HBSC}_i)$ is then distributed in \textbf{S}-$\text{HEP}_{i,\text{ADS},D}$-IRC over $\text{IRC}_{i,j}$, which are the \glspl{IRC} identified based on sensor-relevant condition catalogs, such as CV-HAZOP~\cite{zendel2015cv}. The $\text{IRC}_{i,j}$ occurrence rate under a given $\text{HBSC}_i$ is estimated from field data. Thus, the leaf goal for $D$, \textbf{G}-$\text{HEP}_{i,\text{ADS},D}$-$\text{IRC}_{i,j}$, is to limit the $\text{HEP}_{i,\text{ADS},D}$ rate in each $\text{HBSC}_i$ and $\text{IRC}_{i,j}$, which could be achieved using appropriate test data and other \gls{AI}-component-level measures, as governed by ISO/PAS 8800~\cite{ISO8800}.

The presented template makes several simplifying assumptions, which can be relaxed in a practical application. First, the top-level acceptance criterion does not consider crash severity; this can be addressed by repeating the decomposition for different severities, e.g., $\lambda_{v,\text{S1..3}}$, $\lambda_{v,\text{S2..3}}$, and $\lambda_{v,\text{S3}}$. Further, the template assumes operation within the \gls{ODD}; however, an \gls{ADS} needs to properly recognize and react to out-of-\gls{ODD} situations, such as through a fallback. A complete safety case would also analyse these scenarios. Next, the template assumes $\text{HBSC}_i$ being mutually exclusive. This is not a limitations, since \glspl{HBSC} can always be made disjoint, simply by designating overlaps as separate HBSCs. Further, the template assumes a complete $\text{HBP}_i$ for each $\text{HBSC}_i$; however, in practice, there would be a decomposition of \glspl{HBP} per $\text{HBSC}_i$, such as \gls{UBI} and \gls{UIB} for the sample scenario ``approach to a stationary \gls{POV}.'' This decomposition would be based on scenario catalogs, and it would also require a residual risk estimate due to potentially missed \glspl{HB}. Further, the template states occurrence rates of \glspl{HBSC} and \glspl{IRC} as probabilities without specifying the meaning of a ``trial.'' In practice, these rates would be given as (i) frequencies per hour or kilometer driven for event-based conditions, such as approaches to a stopped vehicle or a stop sign; and (ii) percentages of hour or kilometer driven for conditions that can last arbitrarily long, such as vehicle following~\cite{J2980}. Ultimately, crash rates are given as frequencies per hour or kilometer driven or, equivalently, as a mean number of hours or kilometers driven between crashes. Additionally, the system decomposition in the vehicle- and system-level modules of the template assume mutually exclusive safety-relevant failures of systems and components, respectively, which is sufficient if their rate is kept low to satisfy the acceptance criteria. However, if the criteria satisfaction relies on redundant systems or components, their simultaneous failures need to be modeled explicitly in the decomposition. Lastly, the template is designed for an \emph{ADS} and thus does not deal with human driver behavior. For an \gls{ADAS}, the interaction between the human driver of the \gls{SV} and the system would need to be also considered.

\subsubsection{\gls{MoSAFE} Limitations}

The \gls{HLSM} analysis to derive \glspl{HBP} at the vehicle level (Clause 6) is applicable to any \gls{DAS} architecture and technology, but the \gls{WPP} analysis of the DAS' in the \gls{DSM} at the element level (Clause 7) mainly targets systems that mix conventional software components with \gls{AI}-based components. In particular, the analysis progresses backwards starting from an \gls{HBP} at the \gls{DAS} output and through the models of the conventional components up to the \gls{AI}-based components. It stops at these components, deriving safety-related performance specifications on them as \gls{HEP} occurrence rate limits. The analysis is still applicable to end-to-end optimizable systems that mix neural networks and differentiable variants of classical algorithms. However, the assume-guarantee reasoning espoused by the \gls{WPP} derivation is not applicable to systems that contain mainly neural networks, even recently proposed architectures that strive to achieve modularity~\cite{hu2023_uniad}. This is because some or all of the interfaces in such architectures are latent representations, which are not human-interpretable, and thus it is difficult to impose specifications on them. Whereas some intepretable information can be be extracted from them using suitable decoders, this decoding is incomplete and thus the flow of information and causality among the modules cannot be fully modeled. Future works should explore the development of effective modular reasoning approaches for such end-to-end \gls{AI} systems.

Finally, \gls{MoSAFE} analysis may face limitations in the context of complex scenarios and components, as already discussed. Both in our work and in related works, such as on \gls{RSS}~\cite{shalevshwartz2018formal,Hasuo23RSS}, simple kinematic models and rule-based policies seem to be sufficient to produce adequate \glspl{HLSM} for a wide variety of driving scenarios; however, complex road-user behaviors and complex vehilcle dynamics, especially for high-performance collision avoidance, are likely to defy closed-form solutions and thus require sampling-based approaches in simulation. Similarly, complex component logic may make \gls{WPP} analysis over a \gls{DSM} intractable, in which case the assurance may need to be supplemented by automated black-box testing. Again, our work has demonstrated the feasibility of \gls{WPP} analysis for a sample real-world system, but the limits of such analyses need to be explored in future work.

%% file: related.tex
\section{Related Work}
\label{sec:related}

Probably the most related work targeting \gls{SOTIF} is by Vaicenavicius et al.~\cite{vaicenavicius2020self}, who present an analysis of an automated emergency braking scenario. Although the sample scenario is similar to ours, their work focuses on the statistical analysis of object detection errors that might cause a crash. However, it does not derive temporal specifications of error sequences, and it also acknowledges the toy nature of their illustrative example. In contrast, we have applied \gls{MoSAFE} to four types of \glspl{HB} (i.e., \gls{UBI}, \gls{UIB}, \gls{UA}, and \gls{FTY}) and six different scenarios, and also validated our \gls{UBI} model against the behavior of a real \gls{ADS} in high-fidelity simulator.

Several behavioral safety frameworks, such as \acrfull{RSS}~\cite{shalevshwartz2018formal}, goal-oriented \gls{RSS}~\cite{Hasuo23RSS}, and Safety-Force Field~\cite{SFF19}, define safe behavior of an \gls{SV} while formalizing reasonable behavior of other road users that capture common traffic rules as behavioral contracts and use them as assumptions. In particular, goal-oriented \gls{RSS} formalizes intended \gls{SV} behaviors (aka ``proper responses'') in a range of traffic situations using Hoare quadruples~\cite{Boer97} to allow sequential composition of behavioral contracts in traffic. These quadruples consist of behavior models with pre- and post-condition and invariant specifications, expressed using \glsxtrfull{dFHL}. \gls{MoSAFE} has a different objective from these frameworks, namely to identify and evaluate \glspl{HBP}, i.e., specifications of hazardous deviations from intended behavior (i.e., hazardous deviations from proper response), and then use the identified \glspl{HBP} to identify the corresponding \glspl{HEP} in the \gls{DAS} design. However, the frameworks provide the intended behavior as a starting point for \gls{HBP} identification. Further, \gls{dFHL} could potentially be used to represent the \glspl{HLSM}, \glspl{DSM}, \glspl{HBP}, and \glspl{HEP} and provide a formal basis to develop \gls{WPP} derivation tools based on these models and specifications.

A related family of works applies automated black-box testing in simulation to identify hazardous behaviors of \gls{DAS} (see~\cite{Corso22} for a survey). These approaches use optimization methods, such as genetic algorithms~\cite{Abdessalem18}, importance sampling~\cite{Ding17,Sarkar19}, and reinforcement learning~\cite{Koren18}, to find \gls{DAS} inputs that cause \glspl{HB}, or even estimate their probability (e.g.,~\cite{Ding17,Sarkar19}). As an example, Dreossi et al.~\cite{DreossiDS19} proposed to analyze the \gls{AI}-based components and the conventional ones separately. They run the conventional part of the system with \gls{AI}-based perception replaced by their intended behavior to determine range of low-dimensional inputs $y$ for a given scenario, and then generate images that are consistent with $y$ but cause misperceptions. Finally, they run a falsification tool to search among these images for a sequence to cause a crash. \gls{MoSAFE} is fundamentally different from all these methods. It is model-based, rather than executing the \gls{DAS} as a black-box. It focuses on establishing explicit human-interpretable specifications of hazardous error sequences on a module by module basis. Further, it aligns with the \gls{SOTIF} standard requirements by separating vehicle-level analyses from element-level ones, to allow for reusable \glspl{HBP} that are unaffected by \gls{DAS} implementation. Finally, it establishes a detailed causal model of how hazardous error sequences propagate through the \gls{DAS} in the form of a fault tree with temporal specifications as nodes. The use of models and specifications aids engineers in a deeper understanding of the system that what would be possible with a black-box technique.   

The derivation of \glspl{FT} from software using weakest-precondition reasoning has been explored before~\cite{clarke93}. This related work proposes the use of a weakest-precondition calculus for the programming language at hand to derive software fault trees~\cite{Leveson83}, where the faults could be defective program lines or random hardware errors corrupting program memory. In contrast to the usual approach of using weakest preconditions to characterize safe states or inputs, they are used to specify states or inputs that will result in a given fault. While this related work considers only singular faults as \gls{FT} nodes, our \glspl{FT} use temporal specifications as nodes. There are several temporal extensions of fault trees to capture temporal dependencies among events (see~\cite{Ruijters15} for a comprehensive survey). They include dynamic~\cite{Dugan92} and temporal fault trees~\cite{Palishkar02}, which add different types of temporal gates. None of them consider nodes as temporal specifications, however. We are also not aware of the prior use of implications to represent over-approximation in fault trees.

Finally, in our prior work~\cite{ISCaP}, we introduced the concept of hazardous misperception patterns, which can be viewed as \glspl{HEP} applied to perceptual components, and used them along with \glspl{HBSC} and perception-only (PO) conditions (which correspond to quadrant 1 in Fig.~\ref{fig:steam}) to propose a safety case template for assuring \gls{AI}-based perception as part of a \gls{DAS}. The template, expressed in the goal structuring notation, targets the development of an \acrfull{ISCaP}, which focuses on integrating safety requirements at the system level with perception-component performance requirements at the unit level. \gls{STEAM} generalizes the concepts of misperceptions to \gls{AI} errors and hazardous misperception patterns to hazardous error patterns. Further, \gls{MoSAFE} is complementary to \gls{ISCaP}. In particular, a template similar to \gls{ISCaP} could be used organize the results of \gls{MoSAFE} into an integration safety argument.

%% file: conclusions.tex
\section{Conclusion}

The paper presented \gls{STEAM}, a refinement of the \gls{SOTIF} cause-and-effect model. \gls{STEAM} adds the concept of hazardous error sequences, recognizing the fact that singular errors are often safe, and also adds \glspl{HBP} and \glspl{HEP} as a means to specify classes of hazardous behaviors and hazardous error sequences, respectively. Further, \gls{STEAM} classifies scenario condition as \gls{HB}-sensitive or insensitive and input-relevant or irrelevant, which aids the gradual refinement of scenario models for safety analysis, from \glspl{HLSM} to \glspl{DSM}.

Leveraging \gls{STEAM}, \gls{MoSAFE} helps identify \glspl{HBP} and \glspl{HEP} and evaluate their severity and likelihood. As part of Clause 6 analysis, an \gls{HLSM}, which captures the \glspl{HBSC} and the intended driving policy for the modeled scenario is developed and instrumented to inject \gls{HB} sequences. The instrumented \gls{HLSM} is then used to identify \glspl{HBP} of different severities. The \gls{SOTIF} acceptance targets can then be expressed as upper bounds on the occurrence rate of \glspl{HBP} of different severities.

As part of Clause 7 analysis, a \gls{DSM} and the \glspl{HBP} are used to identify and evaluate \glspl{HEP}. The \gls{DSM} is developed by refining the \gls{HLSM} with \glspl{IRC} and the scenario-relevant \gls{DAS} design and instrumenting it for injecting error sequences. The instrumented \gls{DSM} is then used to identify \glspl{HEP} that cause \glspl{HBP}. The \glspl{HEP} are derived from the \glspl{HBP} as \glspl{WPP}, and the causal links among them are captured in an \gls{FT} with patterns as nodes. Conservative over-approximations of \glspl{WPP} are applied as needed to simplify the analysis and captured in the \gls{FT} as implication arrows. Using the \gls{FT}, the \gls{SOTIF} acceptance targets are then translated into upper bounds on the occurrence rate of the corresponding \glspl{HBP}. This way, safety requirements on the performance of \gls{AI}-based components are established as the upper bounds on the HEP occurrence rates.

The key benefit of \gls{MoSAFE} is its modular and rigorous nature, establishing \glspl{HEP} for each scenario-relevant element in the design and their causal links to \glspl{HBP} in a systematic way. The \gls{WPP} derivation is an instance of a formal assume-guarantee reasoning and thus provides guarantees---under the assumptions made---that are not achievable with simulation testing. Although simulation testing is still needed to validate the models used in \gls{WPP} derivation, it can be more targeted and thus more efficient than without model guidance. Safety engineers are already comfortable with the use of \glspl{FT}, and the \glspl{FT} capturing the causal links among the \glspl{HBP} and \glspl{HEP} can be used as part of a safety case. Finally, the use of models and specifications aids the engineers to acquire a deeper understanding of the system and its potentially hazardous behaviors than what would be possible with black-box testing in simulation. 

\gls{MoSAFE} is subject to several challenges and limitations, which point to exciting directions for future research. First, \gls{MoSAFE} is subject to the same challenges of model creation and maintenance as any other model-based approach. There is an additional effort required to create and maintain models and the quality of the \gls{MoSAFE} results is limited by the adequacy of models. This challenge creates opportunities to research and develop abstraction and slicing tools to help derive scenario-specific models from complete system designs or implementations or both. Additionally, the models need to be validated against implementations, which can be aided by automated black-box testing in simulation, including falsification tools to find counterexamples. Such tools, in combination with adversarial testing of \gls{AI}-based components, could also be used to reduce the residual risk as part of \gls{SOTIF} Clause 11. Second, \gls{WPP} derivation can be challenging in the face of complex decision logic, and the \glspl{WPP} themselves may become complex. Again, this creates an opportunity to research and develop tools to support \gls{WPP} derivation. These could be adaptations of theorem provers as used in proving program correctness, especially for hybrid systems specifications, such as \gls{dFHL}~\cite{Hasuo23RSS}. Another direction is to combine automated black-box testing, such as falsification, with temporal specification mining~\cite{Bartocci22}. The \gls{WPP} derivation should support over-approximation to balance complexity and precision. Furthermore, the tooling should also support \gls{WPP}-based derivation of \glspl{FT}. In particular, multi-failure analyses can be complex and will require adequate tool support. Further refinements of temporal \gls{FT} notations may also be needed. Another opportunities for future research is to model and analyze a wider set of scenarios and \glspl{HB} at the vehicle level, potentially building on other safety frameworks such as goal-oriented \gls{RSS}~\cite{Hasuo23RSS}, which could lead to standard \glspl{HLSM} and \glspl{HBP} that are reusable across the \gls{DAS} industry.

Finally, whereas \gls{MoSAFE} currently targets \gls{DAS} designs that mix conventional and \gls{AI}-based components, future work should address the emerging ``modular'' end-to-end \gls{AI} architectures (e.g.,~\cite{hu2023_uniad}). This means both improving the modularity of these architectures and developing effective assume-guarantee reasoning techniques for them. These techniques will likely be probabilistic and may include causal model learning.

%% file: paper.bbl
\begin{thebibliography}{10}

\bibitem{ISO21448}
International Organization for Standardization, {\em {ISO 21448: Road Vehicles
  -- Safety of the Intended Functionality}}, 2021.

\bibitem{J3016}
SAE, {\em Taxonomy and Definitions for Terms Related to Driving Automation
  Systems for On-Road Motor Vehicles}, 2021.
\newblock J3016\_202104.

\bibitem{ISO26262}
International Organization for Standardization, {\em {ISO 26262: Road Vehicles
  -- Functional Safety}}, 2018.
\newblock 2\textsuperscript{nd} edition.

\bibitem{salay2018using}
R.~Salay and K.~Czarnecki, ``Using machine learning safely in automotive
  software: An assessment and adaption of software process requirements in
  {ISO} 26262,'' {\em arXiv preprint arXiv:1808.01614}, 2018.

\bibitem{J3131-202203}
SAE, {\em Definitions for Terms Related to Automated Driving Systems Reference
  Architecture}, 2022.
\newblock J3131\_202203.

\bibitem{HAZOP}
Ministry of Defence, {\em {Defence Standard 00-58 HAZOP Studies on Items
  Containing Programmable Electronics}}, 2000.

\bibitem{ISO34502}
International Organization for Standardization, {\em {ISO 34502: Road Vehicles
  -- Engineering framework and process of scenario-based safety evaluation}},
  2022.

\bibitem{jurewicz16impact}
C.~Jurewicz, A.~Sobhani, J.~Woolley, J.~Dutschke, and B.~Corben, ``Exploration
  of vehicle impact speed---injury severity relationships for application in
  safer road design,'' {\em Transportation Research Procedia}, vol.~14,
  pp.~4247--4256, 2016.

\bibitem{Maler04}
O.~Maler and D.~Nickovic, ``Monitoring temporal properties of continuous
  signals,'' in {\em Formal Techniques, Modelling and Analysis of Timed and
  Fault-Tolerant Systems} (Y.~Lakhnech and S.~Yovine, eds.), (Berlin,
  Heidelberg), pp.~152--166, Springer Berlin Heidelberg, 2004.

\bibitem{J2980}
SAE, {\em Considerations for {ISO 26262} ASIL Hazard Classification}, 2018.
\newblock J2980\_201804.

\bibitem{krampe20injury}
J.~Krampe and M.~Junge, ``Injury severity for hazard \& risk analyses:
  Calculation of {ISO 26262} {S-parameter} values from real-world crash data,''
  {\em Accident Analysis \& Prevention}, vol.~138, 2020.

\bibitem{bonnett2001stiffness}
G.~M. Bonnett, ``Stiffness coefficients---energy and damage,'' tech. rep.,
  REC-TEC LLC., 2001.
\newblock \url{http://www.rec-tec.com/Energy%20and%20Damage.html}.

\bibitem{ISCaP}
R.~Salay, K.~Czarnecki, H.~Kuwajima, H.~Yasuoka, V.~Abdelzad, C.~Huang,
  M.~Kahn, V.~D. Nguyen, and T.~Nakae, ``The missing link: Developing a safety
  case for perception components in automated driving,'' {\em SAE International
  Journal of Advances and Current Practices in Mobility}, vol.~5, pp.~567--579,
  mar 2022.

\bibitem{ISO34503}
International Organization for Standardization, {\em {ISO 34503: Road Vehicles
  -- Test scenarios for automated driving systems -- Specification for
  operational design domain}}, 2023.

\bibitem{dijkstra75}
E.~W. Dijkstra, ``Guarded commands, nondeterminacy and formal derivation of
  programs,'' {\em Commun. ACM}, vol.~18, p.~453–457, aug 1975.

\bibitem{Pearl09}
J.~Pearl, {\em Causality: Models, Reasoning and Inference}.
\newblock USA: Cambridge University Press, 2nd~ed., 2009.

\bibitem{Ruijters15}
E.~Ruijters and M.~Stoelinga, ``Fault tree analysis: A survey of the
  state-of-the-art in modeling, analysis and tools,'' {\em Computer science
  review}, vol.~15-16, pp.~29--62, May 2015.
\newblock This is the journal published version of technical report
  http://eprints.eemcs.utwente.nl/25404/.

\bibitem{wisedrive}
K.~Czarnecki, ``{Operational World Model Ontology for Automated Driving Systems
  (Part 1 and 2)},'' 2018.
\newblock Waterloo Intelligent Systems Engineering (WISE) Lab,\\Part 1:
  \url{http://dx.doi.org/10.13140/RG.2.2.15521.30568},\\Part 2:
  \url{http://dx.doi.org/10.13140/RG.2.2.11327.00165}.

\bibitem{deGelder19}
E.~de~Gelder, A.~K. Saberi, and H.~Elrofai, ``A method for scenario risk
  quantification for automated driving systems,'' in {\em 26th International
  Technical Conference on the Enhanced Safety of Vehicles (ESV)}, 2019.

\bibitem{XAIforAD}
S.~Atakishiyev, M.~Salameh, H.~Yao, and R.~Goebel, ``Explainable artificial
  intelligence for autonomous driving: {A} comprehensive overview and field
  guide for future research directions,'' {\em CoRR}, vol.~abs/2112.11561,
  2021.

\bibitem{ISO8800}
International Organization for Standardization, {\em {ISO/AWI PAS 8800: Road
  Vehicles -- Safety and Artificial Intelligence}}, 2024.

\bibitem{PURSS}
R.~Salay, K.~Czarnecki, I.~Alvarez, M.~S. Elli, S.~Sedwards, and J.~Weast,
  ``{PURSS}: Towards perceptual uncertainty aware responsibility sensitive
  safety with {ML},'' in {\em AAAI Workshop on Artificial Intelligence Safety
  (SafeAI)}, (New York), CEUR, CEUR, 2020.

\bibitem{Kobayashi21}
T.~Kobayashi, R.~Salay, I.~Hasuo, K.~Czarnecki, F.~Ishikawa, and S.-y.
  Katsumata, ``Robustifying controller specifications of cyber-physical systems
  against perceptual uncertainty,'' in {\em NASA Formal Methods: 13th
  International Symposium, NFM 2021, Virtual Event, May 24–28, 2021,
  Proceedings}, (Berlin, Heidelberg), p.~198–213, Springer-Verlag, 2021.

\bibitem{shalevshwartz2018formal}
S.~Shalev-Shwartz, S.~Shammah, and A.~Shashua, ``On a formal model of safe and
  scalable self-driving cars,'' 2018.
\newblock \textit{arXiv preprint: 1708.06374}.

\bibitem{Hasuo23RSS}
I.~Hasuo, C.~Eberhart, J.~Haydon, J.~Dubut, R.~Bohrer, T.~Kobayashi,
  S.~Pruekprasert, X.-Y. Zhang, E.~A. Pallas, A.~Yamada, K.~Suenaga,
  F.~Ishikawa, K.~Kamijo, Y.~Shinya, and T.~Suetomi, ``Goal-aware {RSS} for
  complex scenarios via program logic,'' {\em IEEE Transactions on Intelligent
  Vehicles}, vol.~8, no.~4, pp.~3040--3072, 2023.

\bibitem{FTA}
NASA, {\em Fault Tree Handbook with Aerospace Fault Tree Handbook with
  Aerospace Applications Applications}, 2002.
\newblock version 1.1.

\bibitem{koopmantoward}
P.~Koopman and M.~Wagner, ``Toward a framework for highly automated vehicle
  safety validation,'' 2018.

\bibitem{Antkiewicz20}
M.~Antkiewicz, M.~Kahn, M.~Ala, K.~Czarnecki, P.~Wells, A.~Acharya, and
  S.~Beiker, ``Modes of automated driving system scenario testing: Experience
  report and recommendations,'' {\em SAE International Journal of Advances and
  Current Practices in Mobility}, vol.~2, pp.~2248--2266, apr 2020.

\bibitem{VanGennip18}
{Van Gennip, Matthew}, ``Vehicle dynamic modelling and parameter identification
  for an autonomous vehicle,'' Master's thesis, University of Waterloo, 2018.
\newblock \url{http://hdl.handle.net/10012/14260}.

\bibitem{Hosking18}
{Hosking, Bryce Antony}, ``Modelling and model predictive control of
  power-split hybrid powertrains for self-driving vehicles,'' Master's thesis,
  University of Waterloo, 2018.
\newblock \url{http://hdl.handle.net/10012/14094}.

\bibitem{Corso22}
A.~Corso, R.~Moss, M.~Koren, R.~Lee, and M.~Kochenderfer, ``A survey of
  algorithms for black-box safety validation of cyber-physical systems,'' {\em
  J. Artif. Int. Res.}, vol.~72, p.~377–428, jan 2022.

\bibitem{NSG21}
J.~Ost, F.~Mannan, N.~Thuerey, J.~Knodt, and F.~Heide, ``Neural scene graphs
  for dynamic scenes,'' in {\em 2021 IEEE/CVF Conference on Computer Vision and
  Pattern Recognition (CVPR)}, pp.~2855--2864, 2021.

\bibitem{Jha19}
S.~Jha, S.~Banerjee, T.~Tsai, S.~K.~S. Hari, M.~B. Sullivan, Z.~T. Kalbarczyk,
  S.~W. Keckler, and R.~K. Iyer, ``{ML}-based fault injection for autonomous
  vehicles: A case for bayesian fault injection,'' in {\em 2019 49th Annual
  IEEE/IFIP International Conference on Dependable Systems and Networks (DSN)},
  pp.~112--124, 2019.

\bibitem{shalev2016sample}
S.~Shalev-Shwartz and A.~Shashua, ``On the sample complexity of end-to-end
  training vs. semantic abstraction training,'' {\em arXiv preprint
  arXiv:1604.06915}, 2016.

\bibitem{Bartocci22}
E.~Bartocci, C.~Mateis, E.~Nesterini, and D.~Nickovic, ``Survey on mining
  signal temporal logic specifications,'' {\em Information and Computation},
  vol.~289, p.~104957, 2022.

\bibitem{ISO5083}
International Organization for Standardization, {\em {ISO/AWI TS 5083: Road
  vehicles -- Safety for automated driving systems -- Design, verification and
  validation}}, 2024.

\bibitem{rushby2015interpretation}
J.~Rushby, ``The interpretation and evaluation of assurance cases,'' {\em Comp.
  Science Laboratory, SRI International, Tech. Rep. SRI-CSL-15-01}, 2015.

\bibitem{GSNstandard}
{SCSC Assurance Case Working Group}, ``Goal structuring notation community
  standard version 3,'' {GSN} Community Standard {SCSC-141C}, Safety-Critical
  Systems Club, CA, USA, 2021.

\bibitem{EN50126}
European Committee for Electrotechnical Standardization (CENELEC), {\em {EN
  50126: Railway Applications. The Specification and Demonstration of
  Reliability, Availability, Maintainability and Safety (RAMS) Generic RAMS
  Process, Part 1}}, 2017.

\bibitem{najm2007pre}
W.~G. Najm, J.~D. Smith, M.~Yanagisawa, {\em et~al.}, ``Pre-crash scenario
  typology for crash avoidance research,'' tech. rep., United States. National
  Highway Traffic Safety Administration, 2007.

\bibitem{safetypool}
{Deepen AI and WMG University of Warwick}, ``Safety pool.''
  \url{https://www.safetypool.ai}.
\newblock Accessed: Jan. 6, 2024 [Online].

\bibitem{zendel2015cv}
O.~Zendel, M.~Murschitz, M.~Humenberger, and W.~Herzner, ``{CV-HAZOP}:
  Introducing test data validation for computer vision,'' in {\em Proceedings
  of the IEEE International Conference on Computer Vision}, pp.~2066--2074,
  2015.

\bibitem{hu2023_uniad}
Y.~Hu, J.~Yang, L.~Chen, K.~Li, C.~Sima, X.~Zhu, S.~Chai, S.~Du, T.~Lin,
  W.~Wang, L.~Lu, X.~Jia, Q.~Liu, J.~Dai, Y.~Qiao, and H.~Li,
  ``Planning-oriented autonomous driving,'' in {\em Proceedings of the IEEE/CVF
  Conference on Computer Vision and Pattern Recognition}, 2023.

\bibitem{vaicenavicius2020self}
J.~Vaicenavicius, T.~Wiklund, A.~Grigaite, A.~Kalkauskas, I.~Vysniauskas, and
  S.~D. Keen, ``Self-driving car safety quantification via component-level
  analysis,'' {\em SAE International Journal of Connected and Automated
  Vehicles}, vol.~4, pp.~35--45, mar 2021.

\bibitem{SFF19}
D.~Nistér, H.-L. Lee, J.~Ng, and Y.~Wang, ``The safety force field,'' tech.
  rep., NVIDIA, 2019.

\bibitem{Boer97}
F.~S. de~Boer, U.~Hannemann, and W.~P. de~Roever, ``Hoare-style compositional
  proof systems for reactive shared variable concurrency,'' in {\em Foundations
  of Software Technology and Theoretical Computer Science} (S.~Ramesh and
  G.~Sivakumar, eds.), (Berlin, Heidelberg), pp.~267--283, Springer Berlin
  Heidelberg, 1997.

\bibitem{Abdessalem18}
R.~Ben~Abdessalem, S.~Nejati, L.~C.~Briand, and T.~Stifter, ``Testing
  vision-based control systems using learnable evolutionary algorithms,'' in
  {\em 2018 IEEE/ACM 40th International Conference on Software Engineering
  (ICSE)}, pp.~1016--1026, 2018.

\bibitem{Ding17}
D.~Zhao, H.~Lam, H.~Peng, S.~Bao, D.~J. LeBlanc, K.~Nobukawa, and C.~S. Pan,
  ``Accelerated evaluation of automated vehicles safety in lane-change
  scenarios based on importance sampling techniques,'' {\em IEEE Transactions
  on Intelligent Transportation Systems}, vol.~18, no.~3, pp.~595--607, 2017.

\bibitem{Sarkar19}
A.~Sarkar and K.~Czamecki, ``A behavior driven approach for sampling rare event
  situations for autonomous vehicles,'' in {\em 2019 IEEE/RSJ International
  Conference on Intelligent Robots and Systems (IROS)}, pp.~6407--6414, 2019.

\bibitem{Koren18}
M.~Koren, S.~Alsaif, R.~Lee, and M.~J. Kochenderfer, ``Adaptive stress testing
  for autonomous vehicles,'' in {\em 2018 IEEE Intelligent Vehicles Symposium
  (IV)}, pp.~1--7, 2018.

\bibitem{DreossiDS19}
T.~Dreossi, A.~Donz{\'{e}}, and S.~A. Seshia, ``Compositional falsification of
  cyber-physical systems with machine learning components,'' {\em J. Autom.
  Reason.}, vol.~63, no.~4, pp.~1031--1053, 2019.

\bibitem{clarke93}
S.~J. Clarke and J.~A. McDermid, ``Software fault-trees and weakest
  preconditions - a comparison and analysis,'' {\em Software Engineering
  Journal}, vol.~8, pp.~225--236, July 1993.

\bibitem{Leveson83}
N.~G. Leveson and P.~R. Harvey, ``Software fault tree analysis,'' {\em Journal
  of Systems and Software}, vol.~3, no.~2, pp.~173--181, 1983.

\bibitem{Dugan92}
J.~Dugan, S.~Bavuso, and M.~Boyd, ``Dynamic fault-tree models for
  fault-tolerant computer systems,'' {\em IEEE Transactions on Reliability},
  vol.~41, no.~3, pp.~363--377, 1992.

\bibitem{Palishkar02}
G.~K. Palshikar, ``Temporal fault trees,'' {\em Information and Software
  Technology}, vol.~44, no.~3, pp.~137--150, 2002.

\end{thebibliography}
